\documentclass{article}

\pdfoutput=1

\usepackage[preprint]{neurips_2022}

\usepackage{natbib}
\setcitestyle{numbers,square}

\usepackage[utf8]{inputenc} 
\usepackage[T1]{fontenc}    
\usepackage{hyperref}       
\usepackage{url}            
\usepackage{booktabs}       
\usepackage{amsfonts}       
\usepackage{nicefrac}       
\usepackage{microtype}      
\usepackage{xcolor}         

\usepackage{enumitem}
\usepackage{amsmath}
\usepackage{amssymb}
\usepackage{mathtools}
\usepackage{amsthm}
\usepackage{bm}
\usepackage{subfigure}
\usepackage{multirow}
\usepackage{float}

\newcommand{\PD}{{\mathbf{v}}}
\newcommand{\CT}{{\mathbf{M}}}

\newcommand{\M}{SHoP}
\newcommand{\f}{\frac}
\newcommand{\p}{\partial}
\newcommand{\B}{\mathbf}

\title{\M: A Deep Learning Framework for Solving High-order Partial Differential Equations}

\author{
    Tingxiong Xiao\textsuperscript{1}~~~~~~~~ Runzhao Yang\textsuperscript{1}~~~~~~~~ Yuxiao Cheng\textsuperscript{1} \\
    \textbf{Jinli Suo\textsuperscript{12}\footnotemark[1]~~~~~~~~ Qionghai Dai\textsuperscript{12}\footnotemark[1]\thanks{Corresponding author}}\\
    \textsuperscript{1}Department of Automation, Tsinghua University\\
    \textsuperscript{2}Institute for Brain and Cognitive Science, Tsinghua University (THUIBCS)\\
    \texttt{\{xtx22,yangrz20,cyx22\}@mails.tsinghua.edu.cn} \\
    \texttt{\{jlsuo,qhdai\}@tsinghua.edu.cn} \\
}

\begin{document}

\maketitle

\begin{abstract}
Solving partial differential equations (PDEs) has been a fundamental problem in computational science and of wide applications for both scientific and engineering research. Due to its universal approximation property, neural network is widely used to approximate the solutions of PDEs. However, existing works are incapable of solving high-order PDEs due to insufficient calculation accuracy of higher-order derivatives, and the final network is a black box without explicit explanation. To address these issues, we propose a deep learning framework to solve high-order PDEs, named \M. Specifically, we derive the high-order derivative rule for neural network, to get the derivatives quickly and accurately; moreover, we expand the network into a Taylor series, providing an explicit solution for the PDEs. We conduct experimental validations four high-order PDEs with different dimensions, showing that we can solve high-order PDEs efficiently and accurately.
\end{abstract}

\section{Introduction}
Partial differential equations (PDEs) are used to describe the basic rules underlying complex processes in both scientific and engineering fields, and researchers have devoted lots of efforts to developing algorithms searching for their numerical solutions. Conventional finite difference methods become infeasible for high scale PDEs due to the difficulty in constructing mesh explicitly. In the past two decades, with the rapid development of deep neural networks, ones began to utilize its universal approximating capability \cite{cybenko1989approximation,hornik1989multilayer,chen1995universal,leshno1993multilayer} to fit the solutions of PDEs, including its differential operator and constraints (the initial condition and boundary conditions), if any. Being a mesh-free approach, deep learning based solvers can circumvent the grand challenges in terms of memory and searching time when tackling high scale PDE.


The pioneering work utilizing neural networks to solve PDEs can date back to 1990s, when Dissanayake et al. proposed to use an MLP to find PDE's numerical solution  \cite{dissanayake1994neural} with easy implementation and high running efficiency, but with low proximity. 
With the rapid development in deep learning, PDE solvers based on deep neural networks are gaining momentum recently. 
PDE-FIND\cite{pde-find} proposes a sparse regression method capable of discovering the governing partial differential equation(s) of a given system by measured time series in the spatial domain, and demonstrates its computation efficiency, robustness, and applicability on a variety of canonical problems spanning a number of scientific domains. 
Later, Deep Ritz Method\cite{yu2018deep} is designed to solve PDEs via approximating its analytical solution using a deep neural network, using the estimators in Ritz method to train the network and obtain the approximate solution. To solve high-dimensional PDEs, Sirignano et al. draw inspirations from Galerkin methods and proposed Deep Galerkin Method (DGM)\cite{sirignano2018dgm}, with the solution approximated by a neural network instead of a linear combination of basis functions. 
In 2019, Raissi et al. attempt to incorporate the principled physical constraints into learning of the deep neural networks, named Physics-informed neural networks (PINNs)\cite{raissi2019physics}, targeting for reducing demanding training data, raising robustness and accelerate convergence. This approach can both search for data-driven solution and conduct data-driven discovery of partial differential equations, and fire up a series of working for further improvement. For example, DeepXDE \cite{lu2021deepxde} improves its training efficiency using  a new residual-based adaptive refinement (RAR) method and provides a Python library for PINNs as an educational and research tool; there are also some work\cite{wang2021understanding,liu2021dual,xiang2021self} studying the composite loss functions in the training process to accelerate the convergence or improve the final accuracy. 
Some  other researchers \cite{meng2020ppinn,moseley2021finite,lyu2022mim} adapt the domain decomposition technology in traditional PDEs methods to realize parallel operations in time and space, or decomposes the order of derivatives to reduce the complexity and difficulty of single network learning. 
In the most recent years, Lu et al. design DeepONet\cite{lu2021learning}, a new network structure consisting of a branch net and a trunk net to encode the discrete input function space and output functions. Under this new architecture, one can learn various explicit operators, such as integrals and fractional Laplacians, as 
well as implicit operators that represent deterministic and stochastic differential equations. 

In spite of the striding progress, existing work is still at early stage and are faced with at least two challenges. Firstly, they are incapable of handling high-order PDEs, which is an important branch in PDE with wide applications.   
Most of above works adapt automatic differentiation module, like Autograd\cite{paszke2017automatic} to calculate the derivatives under the current input, and use some optimization algorithms like Adamax\cite{kingma2014adam} to optimize the network parameters. Autograd uses computation graph to record the intermediate process, and back-propagate to calculate the derivatives based on the existing computation graphs. However, as the order of derivative increase to a certain extent, an amount of calculation graphs need to be created and thus the calculation becomes intractable, in terms of both explosive growth of memory and inference time.  
On the other hand, as we all know, most neural networks are black box and the lacking interpretability hampers its practical applications, even with excellent performance. Designing algorithms with explicit explanations of the differentiation operators is key for pushing forward the deep-neural-network-based solvers towards real applications.  

To solve the above two issues, we propose a deep learning framework to solve high-order PDEs, named \M, being able to solve high-order PDEs in explicit manner. Theoretically, it is proved that when the activation function is infinitely differentiable, the neural network is equivalent to its Taylor series, and when the network parameters meet certain distribution rules the Taylor series converges. Unlike the computation graph, our method gives an explicit formula for calculating the first $n$-order derivatives, which brings two-fold benefits. Firstly, after calculating the transformation matrix, we can get all the derivatives in just one step, more quickly and accurately than computation graph, and can greatly save memory resources. Secondly, once the network trained, we can expand the black-box network into an explicit expression of Taylor series if needed. 
We tested \M~ on four types of PDEs, and experimentally show that \M~ can solve the equation efficiently and accurately. 

\vspace{3mm}
To summarize, the technical contributions are as follows:
\begin{itemize}[leftmargin=*]
    \setlength{\parskip}{0pt}
\item We propose the high-order derivative rule of neural network to calculate the derivatives quickly and accurately.
\item We solve the high-order PDEs under the new derivative rule, via calculating the high-order derivatives in just one step, with higher accuracy, higher speed and less memory consumption than conventional computation graph. 
\item We propose to expand a neural network into Taylor series, providing an explicit explanation for the neural network fitting the PDE solution.
\item We prove the equivalence between a neural network and its Taylor series, and analyze its convergence condition. 
\end{itemize}

\section{High-order derivatives of neural network for solving PDEs}\label{High-order derivatives}
As known, we can describe the underlying solution of a PDE with a deep neural network and optimize the network parameters in a data driven manner. Mathematically, the key module of the solver is to calculate the derivatives of output with respect to the input, and here we propose an efficient method to get the high-order derivatives.

\subsection{High-order derivatives of composite function}\label{High-order derivatives of composite function}
Considering a composite function $f(g(x))$, with $g(x)$ and $f(z)$ being $n$-order differentiable at $x0$ and $z0=g(x0)$ respectively. $\frac{\partial^k g}{\partial x^k}|_{x0}$ and $\frac{\partial^k f}{\partial g^k}|_{z0}$ are the $k$-order derivative of $g(x)$ at $x0$ and of $f(z)$ at $z0$. According to the chain rule, we can calculate the first three terms of $f(g(x))$'s $n$-order derivatives as
\begin{equation}  
\begin{split}
\left\{  
\begin{array}{lc}
    \frac{\partial f}{\partial x}
    =\frac{\partial g}{\partial x}\frac{\partial f}{\partial g},  \\    
    \frac{{\partial}^2 f}{\partial x^2}
    =\frac{\partial^2 g}{\partial x^2}\frac{\partial f}{\partial g}+(\frac{\partial g}{\partial x})^2\frac{\partial^2 f}{\partial g^2}, \\
    \frac{{\partial}^3 f}{\partial x^3}
    =\frac{\partial^3 g}{\partial x^3}\frac{\partial f}{\partial g}+3\frac{\partial g}{\partial x}\frac{\partial^2 g}{\partial x^2}\frac{\partial^2 f}{\partial g^2}+(\frac{\partial g}{\partial x})^3\frac{\partial^3 f}{\partial g^3}.
\end{array}
\right.  
\label{3-order derivatives}
\end{split}
\end{equation} 
For more terms, we convert $\frac{\partial }{\partial x}\frac{{\partial}^i f}{\partial g^i}$ to $\frac{\partial g}{\partial x}\frac{{\partial}^{i+1} f}{\partial g^{i+1}}$, and 
$\frac{{\partial}^n f}{\partial x^n}$ can be calculated given $\{\frac{\partial^i f}{\partial g^i}, i=1,\ldots,n\}$ and $\{\frac{\partial^i g}{\partial x^i}, i=1,\ldots,n\}$. Then Eq.~(\ref{3-order derivatives}) turns into following matrix form
\begin{equation}
\begin{split}
\left[
    \begin{array}{c}
            \frac{\partial f}{\partial x}  \\
            \vdots \\
            \frac{\partial^n f}{\partial x^n}
    \end{array}
\right]=
\left[
    \begin{array}{cccc}
            \frac{\partial g}{\partial x} & 0 & 0 & 0  \\
             \frac{\partial^2 g}{\partial x^2} & (\frac{\partial g}{\partial x})^2 & 0 & 0 \\
             \frac{\partial^3 g}{\partial x^3} & 3\frac{\partial g}{\partial x}\frac{\partial^2 g}{\partial x^2} & (\frac{\partial g}{\partial x})^3 & 0 \\
            \vdots & \vdots & \vdots & \ddots
    \end{array}
\right]
\left[
    \begin{array}{c}
            \frac{\partial f}{\partial g}  \\
            \vdots \\
            \frac{\partial^n f}{\partial g^n}
    \end{array}
\right],
\end{split}
\label{chain matrix}
\end{equation}
which can be further abbreviated as
\begin{equation}
    \PD^{f,x}=\CT^{g,x}\PD^{f,g}.
\label{abbreviated chain matrix}
\end{equation}
In this equation $\PD^{f,x} \in \mathbb{R}^n$ and $\PD^{f,g} \in \mathbb{R}^n$ are respectively the vectors composed of partial derivatives $\{\frac{\partial^i f}{\partial x^i}\}$ and  $\{\frac{\partial^i f}{\partial g^i}\}$; $\CT^{g,x} \in \mathbb{R}^{n\times n}$ is a transformation matrix composed of $\frac{\partial^i g}{\partial x^i}$ and takes a lower triangular form. So far, the calculation of $f(g(x))$'s $n$-order derivatives turns into the computation of $\CT^{g,x} \in \mathbb{R}^{n\times n}$. The recurrence formula of  $\CT^{g,x}$ is
\begin{equation}  
\left\{  
    \begin{array}{lc}
        \CT^{g,x}_{1,1}=\frac{\partial g}{\partial x}  \\  
        \CT^{g,x}_{i,j}=0, i < j\\  
        \CT^{g,x}_{i+1,j}=\frac{\partial \CT^{g,x}_{i,j}}{\partial x}
         +\frac{\partial g}{\partial x}\CT^{g,x}_{i,j-1},  
    \end{array}  
\right.
\label{recurrence}
\end{equation} 
which explicitly composes the $n$-order chain transformation matrix $\CT^{g,x}$ in Eq.~(\ref{abbreviated chain matrix}). The detailed derivation process of Eq.~(\ref{recurrence}) can be found in Supplementary Materials.

\subsection{High-order derivatives of neural network}

\begin{figure}[t]
\begin{center}
\centerline{\includegraphics[width=\columnwidth]{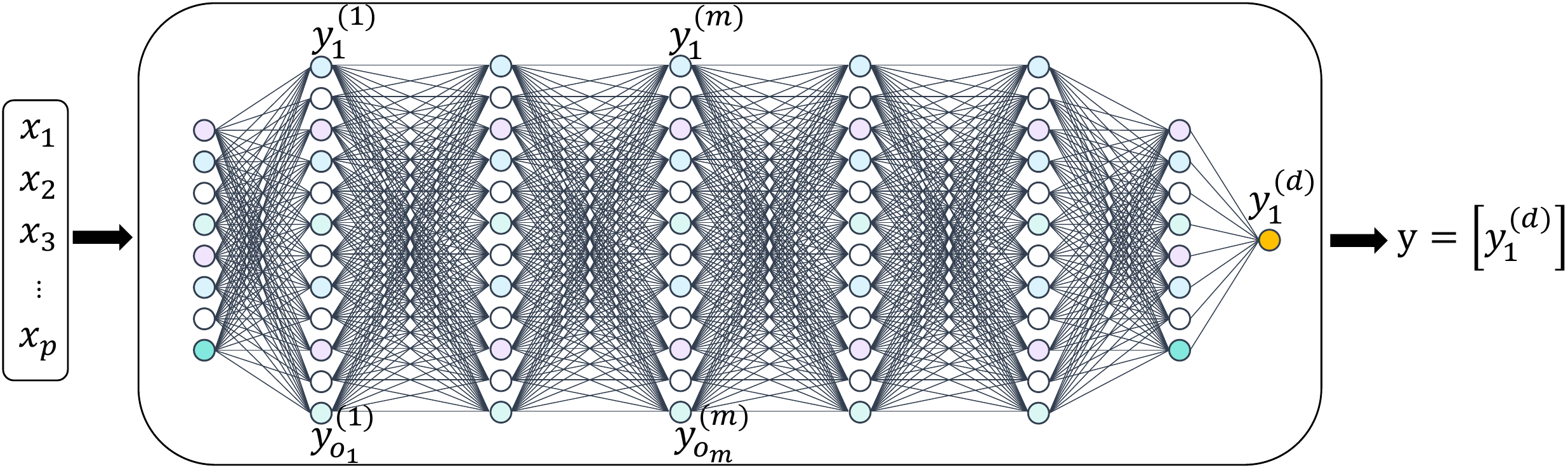}}
\vspace{-1mm}
\caption{The structure of a single-output multilayer perceptron with $d$ layers and $o_m$ nodes in the $m$th layer. The network maps the input $\mathbf x\!=\![x_1\!\cdots\! x_p]^T\in \mathbb{R}^p$ to $\mathbf{y}=\mathbf{y}^{(d)}=\left[y_1^{(d)}\right]\in \mathbb{R}$, with the intermediate output of the $i$-th node in $m$th layer being $y_{i}^{(m)}$.}
\label{network}
\end{center}
\vspace{-3mm}
\end{figure}

Without loss of generality, we take Multilayer Perceptron (MLP) as an example, with the network structure illustrated in Fig.~{\ref{network}}. Denoting the input as $\mathbf{x}=[x_1 \ldots x_p]^T\in \mathbb{R}^p$, the network depth as $d$, the width of $m$th layer as $o_m$, the output of the $i$-th node in $m$th layer as ${y}_{i}^{(m)}$, the linear weighted result of input of the $i$-th node in $m$th layer as ${z}_{i}^{(m)}$, and the final output as $\mathbf{y}=\mathbf{y}^{(d)}=\left[y_1^{(d)}\right]\in \mathbb{R}$, this article aims to calculate $\mathbf{y}$'s $n$-order derivatives with respect to input $\mathbf{x}$. 
The input-output relationship of the MLP can be described explicitly, with the first layer being 
\begin{equation}
\begin{split}
    {y}_{i}^{(1)}=\sigma ({z}_{i}^{(1)})
    =\sigma (\sum_{j=1}^{p}\mathbf{W}_{i,j}^{(1)}{x}_j+\mathbf{b}_i^{(1)}),
    \label{input function}
\end{split} \hspace{1cm}
\end{equation}
and the successive layers as
\begin{equation}
\begin{split}
    {y}_{i}^{(m+1)}\!=\!\sigma ({z}_{i}^{(m+1)})
    \!=\!\sigma (\sum_{j=1}^{o_m}\mathbf{W}_{i,j}^{(m+1)}{y}_j^{(m)}\!+\!\mathbf{b}_i^{(m+1)}).
\end{split}
\label{layer function}
\end{equation}
Then the final output is defined as 
\begin{equation}
    \mathbf{y}=\mathbf{y}^{(d)}=\left[y_1^{(d)}\right], 
\label{output function}  
\end{equation}
in which $\mathbf{W}^{(m+1)} \in \mathbb{R}^{o_{m+1}\times o_{m}}$ is the weight matrix of layer $m+1$, $\mathbf{b}^{(m+1)} \in \mathbb{R}^{o_{m+1}}$ is the bias vector, $\sigma (*)$ is the nonlinear activation function. $\mathbf{y}^{(m)}=\left[y_1^{(m)},\ldots,y_{o_m}^{(m)}\right]^T \in \mathbb{R}^{o_m}$ is the output vector of $m$th layer. 

With above denotations, we induce $\mathbf{y}$'s derivatives  with respect to the input $\mathbf{x}$. 
First, from Eq.~(\ref{output function}) and the definition in Eq.~\ref{abbreviated chain matrix}, we can get a initial partial derivative vector
\begin{equation}
    \PD^{\mathbf{y},{y_1}^{(d)}}=
    \left[
    \begin{array}{cccc}
        \frac{\partial \mathbf{y}}{\partial {{y}_1^{(d)}}} & \frac{\partial^2 \mathbf{y}}{\partial {{y}_1^{(d)}}^2} & \ldots & \frac{\partial^n \mathbf{y}}{\partial {{y}_1^{(d)}}^n}
    \end{array}
    \right]^T=
    \left[
    \begin{array}{cccc}
        1 & 0 & \ldots & 0
    \end{array}
    \right]^T,
\label{eq:inti}
\end{equation}

Taking derivatives over both sides of Eq.~(\ref{layer function}) arrives at
\begin{equation}
    \frac{\partial^k {y}_{i}^{(m+1)}}{\partial {{y}_{j}^{(m)}}^k}={\mathbf{W}_{i,j}^{(m+1)}}^k
    \frac{\partial^k \sigma ({z}_{i}^{(m+1)})}{\partial {{z}_{i}^{(m+1)}}^k}, 
\label{yi yj}
\end{equation}
which form the basic elements of matrix $\CT^{{y}_i^{(m+1)},{y}_j^{(m)}}$ defined in Eq.~(\ref{abbreviated chain matrix}).

Given $\{\frac{\partial^k \mathbf{y}}{\partial {{y}_{i}^{(m+1)}}^k}\}$ and $\{\frac{\partial^k y_i^{(m+1)}}{\partial {{y}_{j}^{(m)}}^k}\}$, according to Eqns.~(\ref{chain matrix})(\ref{abbreviated chain matrix})(\ref{recurrence}), we can calculate $\frac{\partial^k \mathbf{y}}{\partial {{y}_{j}^{(m)}}^k}$ as
\begin{equation}
    \PD^{\mathbf{y},{y}_j^{(m)}}=\sum_{i=1}^{o_{m+1}}
    \CT^{{y}_i^{(m+1)},{y}_j^{(m)}}
    \PD^{\mathbf{y},{y}_i^{(m+1)}}.\label{eq:transfer}
\end{equation} 
By analogy, we can get all the unmixed partial derivatives $\{\frac{\partial^k \mathbf{y}}{\partial {{x}_{j}}^k}, j=1,\ldots,p\}$ by calculating $\PD^{\mathbf{y},{x}_j}$. 

For the mixed partial derivatives, 
we can get $\{\frac{\partial^k \mathbf{y}}{\partial {{y}_{q}^{(1)}}^k}, q=1,\cdots,o_1\}$ and calculate them. For example,
\begin{equation}
    \begin{split}
        \frac{\partial^2 \mathbf{y}}{\partial {x}_1 \partial {x}_2}
        &=\sum_{q=1}^{o_1}\frac{\partial }{\partial {x}_2}
        \left(\frac{\partial {y}_q^{(1)}}{\partial {x}_1}
        \frac{\partial \mathbf{y}}{\partial {y}_q^{(1)}}\right)
        =\sum_{q=1}^{o_1}\left(\frac{\partial^2 {y}_q^{(1)}}{\partial {x}_1\partial {x}_2}
        \frac{\partial \mathbf{y}}{\partial {y}_q^{(1)}}+\frac{\partial {y}_q^{(1)}}{\partial {x}_1}
        \frac{\partial {y}_q^{(1)}}{\partial {x}_2}
        \frac{\partial^2 \mathbf{y}}{\partial {{y}_q^{(1)}}^2}\right) \\
        &=\sum_{q=1}^{o_1}\mathbf{W}_{q,1}^{(1)}\mathbf{W}_{q,2}^{(1)}
        \left(\frac{\partial^2 \sigma({z}_q^{(1)})}{\partial {{z}_q^{(1)}}^2}\frac{\partial \mathbf{y}}{\partial {y}_q^{(1)}}\!+\!{(\frac{\partial \sigma({z}_q^{(1)})}{\partial {z}_q^{(1)}})}^2\!\frac{\partial^2 \mathbf{y}}{\partial {{y}_q^{(1)}}^2}\right).
    \end{split}\label{eq:mix}
\end{equation}
Therefore, during the back-propagation of the neural network, we do not need to calculate the mixed partial derivatives like $\frac{\partial^k \mathbf{y}}{\partial {{y}_{i}^{(m)}}^{k-1}\partial {y}_{j}^{(m)}}$, but only need to calculate $\frac{\partial^k \mathbf{y}}{\partial {{y}_{i}^{(m)}}^k}$ instead. Further according to the chain rule, we can get all the mixed partial derivatives from $\frac{\partial^k \mathbf{y}}{\partial {{y}_{q}^{(1)}}^k}$ in one time.

For faster calculation of the partial derivatives, we convert the above formulas into matrix form, and the detailed formulas can be find in Supplementary Materials.

\section{\M: A deep learning framework to solve high-order PDEs}\label{framework}
After calculating the partial derivatives, we can solve a PDE via designing a sampler and constructing a loss function. 

\subsection{The working flow of \M}

Considering a PDE with $p$ dimensions
\begin{equation}  
\left\{  
    \begin{array}{lc}
    \mathcal{L}u(\mathbf{x})=0, & \mathbf{x}\in \Omega \\
    u(\mathbf{x})=g(\mathbf{x}), & \mathbf{x}\in \partial\Omega
    \end{array}  
\right.  
\end{equation} 
where $\mathbf{x}\in \Omega \subset \mathbb{R}^p$, $\partial\Omega$ is $\Omega$'s boundary, $\mathcal{L}u(\mathbf{x})$ is a combination of derivatives of $u(\mathbf{x})$ with respect to $\mathbf{x}$. We use a neural network $f(\mathbf{x};\theta)$ to approximate $u(\mathbf{x})$ with $\theta$ being the network parameters.  

In terms of the sampler, before network training, we establish discrete coordinates according to $\Omega$ and $\partial \Omega$, and randomly sample the coordinate points according to the preset batch size during training. 

The objective function is defined as 
\begin{equation}
\begin{split}
    J(\mathbf{x};\theta)=\lambda\Vert \mathcal{L}f(\mathbf{x};\theta) \Vert_{\Omega}^2+
    \mu\Vert f(\mathbf{x};\theta)-g(\mathbf{x}) \Vert_{\partial \Omega}^2.
\end{split}
\end{equation} 
Here the two terms are $\emph{l}_2$ norm fitting the partial derivatives over the defining field $\Omega$ and along the boundary $\partial \Omega$ respectively, $\lambda$ and $\mu$ are hyper-parameters balancing two terms in the loss function. 

In each training epoch, after applying forward propagation on the neural network, we calculate a set of partial derivatives with the new derivative method. For example, when $p$=2 and $n$=3, the unmixed partial derivatives include $\left[f,~f_{x_1},~f_{x_2},~f_{x_1x_1}, f_{x_2x_2}, f_{x_1x_1x_1}, f_{x_2x_2x_2}\right]$ with footnote representing the differential order, and the mixed derivatives includes $\left[f,~f_{x_1},f_{x_2},f_{x_1x_1}, f_{x_1x_2}, f_{x_2x_1},f_{x_2x_2},f_{x_1x_1x_1},
f_{x_1x_1x_2},\!f_{x_1x_2x_1},\!f_{x_1x_2x_2},\!f_{x_2x_1x_1},\! f_{x_2x_1x_2},\!f_{x_2x_2x_1},\notag \right. \\ \left.
\!   f_{x_2x_2x_2}\right]$. In implementation, we calculate all these derivatives in one time, and then retrieve the required terms following their indices to compute the fitting error to the loss function.

\subsection{Explicit expression of the PDE solution}
With the first $n$-order derivatives, we can get an explicit Taylor series to approximate the original network locally. The $n$-order Taylor series can be calculated as 
\begin{equation}
\begin{split}
    f(\B x)=f(\mathbf{x}0;\theta) + \sum_{i=1}^{p} \f{\f{\p f(\mathbf{x};\theta)}{\p\B x_i}|_{\B x0}}{1!} \Delta\B x_i+\ldots
    +\sum_{i_1,\ldots,i_n=1}^{p} \f{\f{\p^n f(\mathbf{x};\theta)}{\p\B x_{i_1}\ldots\p\B x_{i_n}}|_{\B x0}}{n!} \Delta\B x_{i_1}\ldots\Delta\B x_{i_n},
\end{split}
\end{equation}
where $\Delta\B x=\B x-\B x0$, $\f{\p^k f(\mathbf{x};\theta)}{\p\B x_{i_1}\ldots\p\B x_{i_k}}|_{\B x0}$ is a $k$-order partial derivative on the reference point $\B x0$. 

Although approximating a PDE's solution with a deep neural network is of high efficiency and accuracy, as we all know, such a black-box model lacks interpretability and hampers its practical applications. After Taylor expansion, one can retrieve the Taylor series out of the black-box explicitly, which can provide us a deeper understanding of the mapping mechanism of the learned neural network. Here we give two potential studies benefiting from such explicit expansion: (i) In a complex process with multiple input, after expanding the governing neural network into Taylor series, one can quantify the contribution of each input. Such explicit description might inspire researchers to analyze the underlying causation mechanism of the target output. 
(ii) We can also bridge the network parameters (mapped to the weights of Taylor series) and the domain expertise, and thus measure the reliability of the deep neural network interpretably and set proper confidence level to the network output. In other words, our expansion facilitates studying the fidelity of the neural network in a more explainable way and advancing its real applications.

\subsection{Analysis of the Taylor polynomial convergence}\label{convergence}
When the activation function is infinitely differentiable, we can calculate all the derivatives and thus the neural network is equivalent to its Taylor series. From Eqns.~(\ref{yi yj})(\ref{eq:mix}), the $k$-order derivatives are related to $\mathbf{W}_{i_1,j_1}^{(m)}\mathbf{W}_{i_2,j_2}^{(m)}\ldots \mathbf{W}_{i_k,j_k}^{(m)}$, which is the continuous multiplication of $k$ weights in $\mathbf{W}^{(m)}$. 
\begin{equation}
\begin{split}
    |\frac{\partial^k\B y}{\partial\B x^k}|\propto |\mathbf{W}_{i,j}^{(m)}|^k.
\end{split}
\end{equation}

When the parameters in $\mathbf W^{(m)}$ is concentrated near 0, higher-order derivatives are more likely to approach 0. When the parameters is located far from 0, higher-order derivatives may become increasingly larger due to continuous addition and multiplication, and thus the Taylor series diverge, i.e., we cannot obtain the Taylor approximate solution.
\begin{equation}
\begin{split}    \lim_{\substack{|\mathbf{W}_{i,j}^{(m)}| \to 0 \\ k \to \infty}} |\frac{\partial^k\B y}{\partial\B x^k}| = 0, ~~~    \lim_{\substack{|\mathbf{W}_{i,j}^{(m)}| >1 \\ k \to \infty}} |\frac{\partial^k\B y}{\partial\B x^k}| = +\infty.
\end{split}
\end{equation}
The above analysis tells that the parameter distribution of each layer has a great influence on the convergence of Taylor expansion. The above rules help imposing constraints on the network parameters during the network training, and can also help designing network structures with high-order Taylor approximation. More theory details can be found in Supplementary Materials.

\begin{figure}[t]
\vspace{-6mm}
\begin{center}
    \subfigure[]{
        \includegraphics[width=0.3\columnwidth]{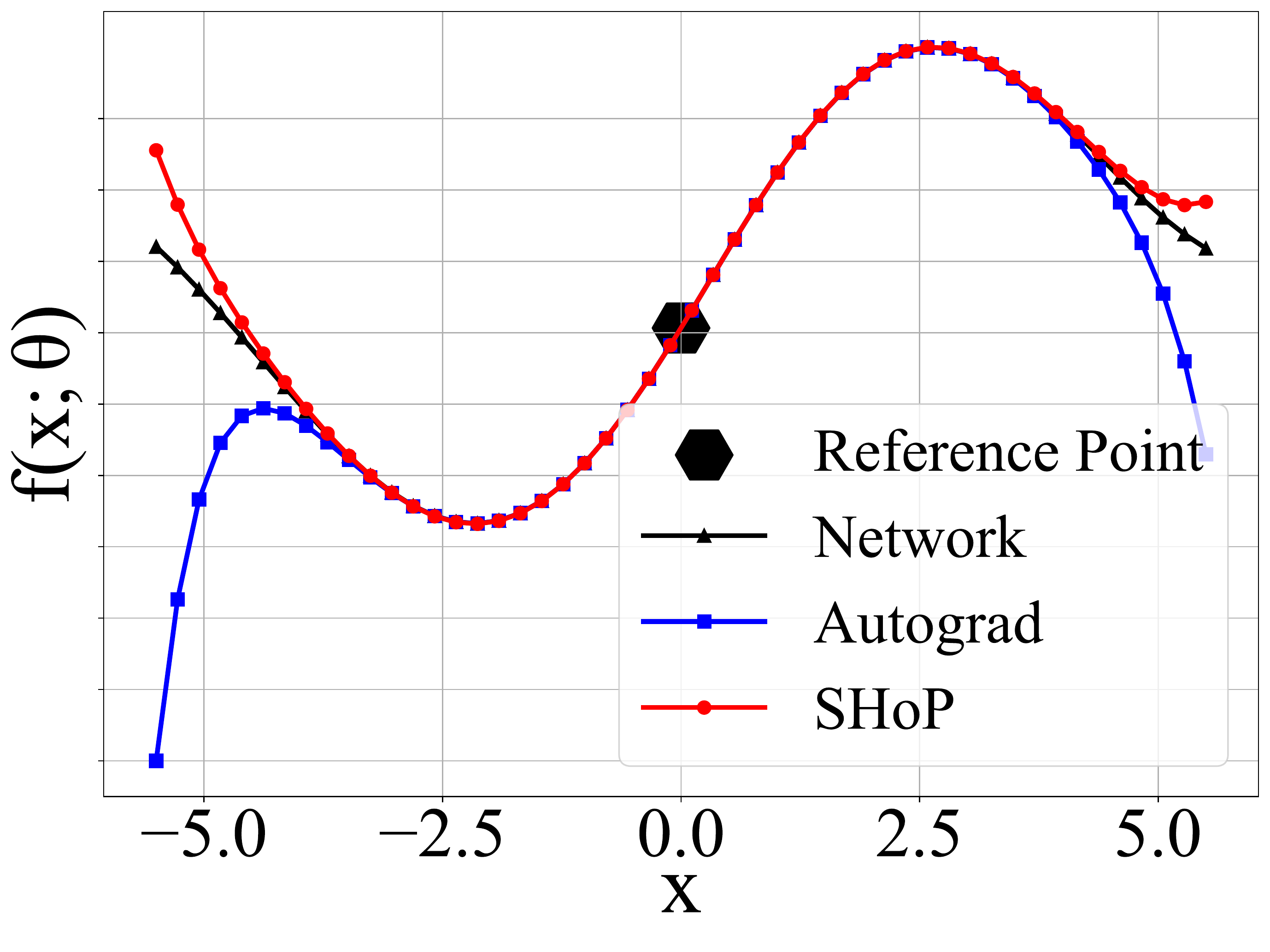}
        \hspace{3mm}
        \includegraphics[width=0.3\columnwidth]{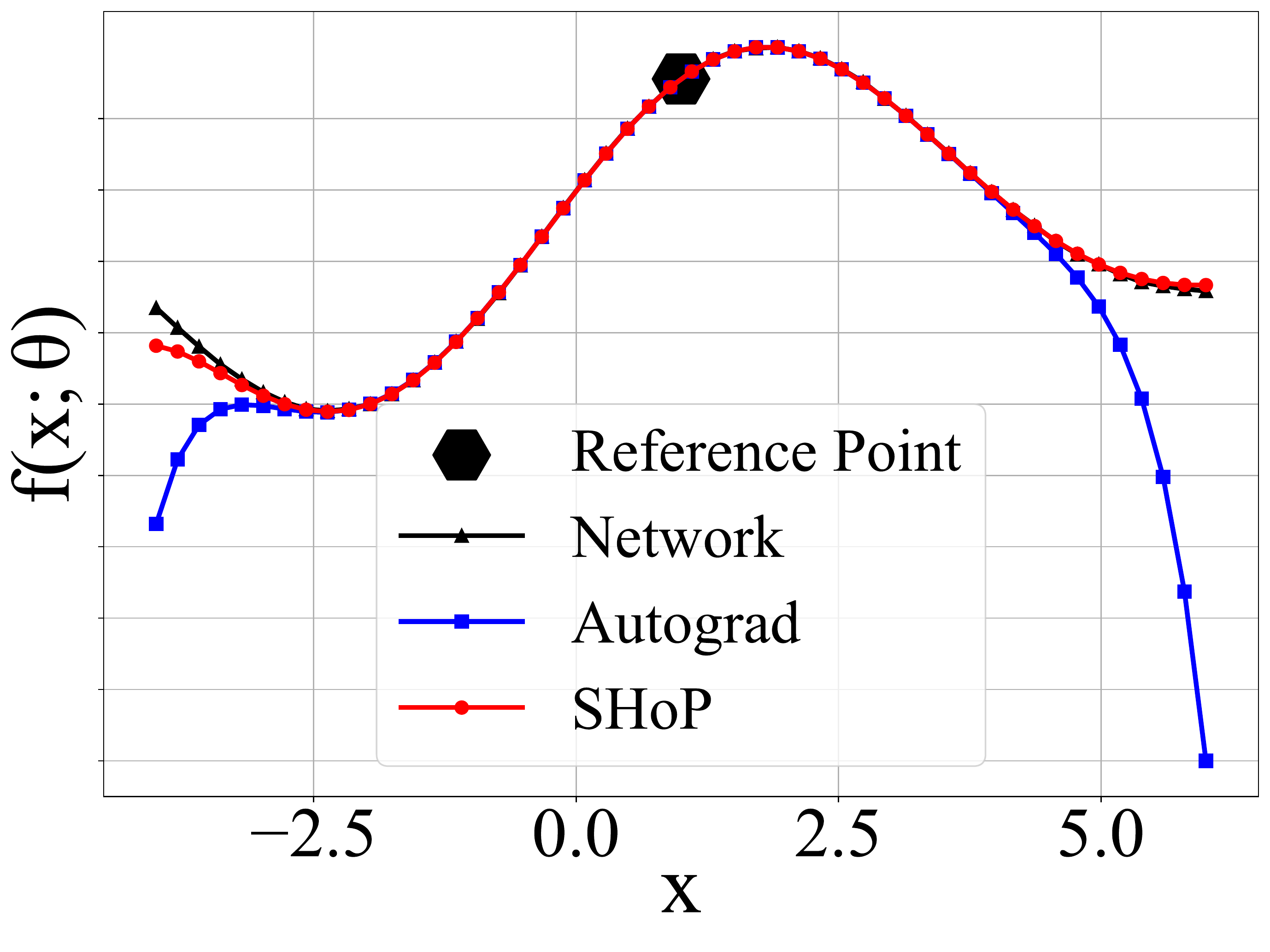}
        \hspace{3mm}
        \includegraphics[width=0.3\columnwidth]{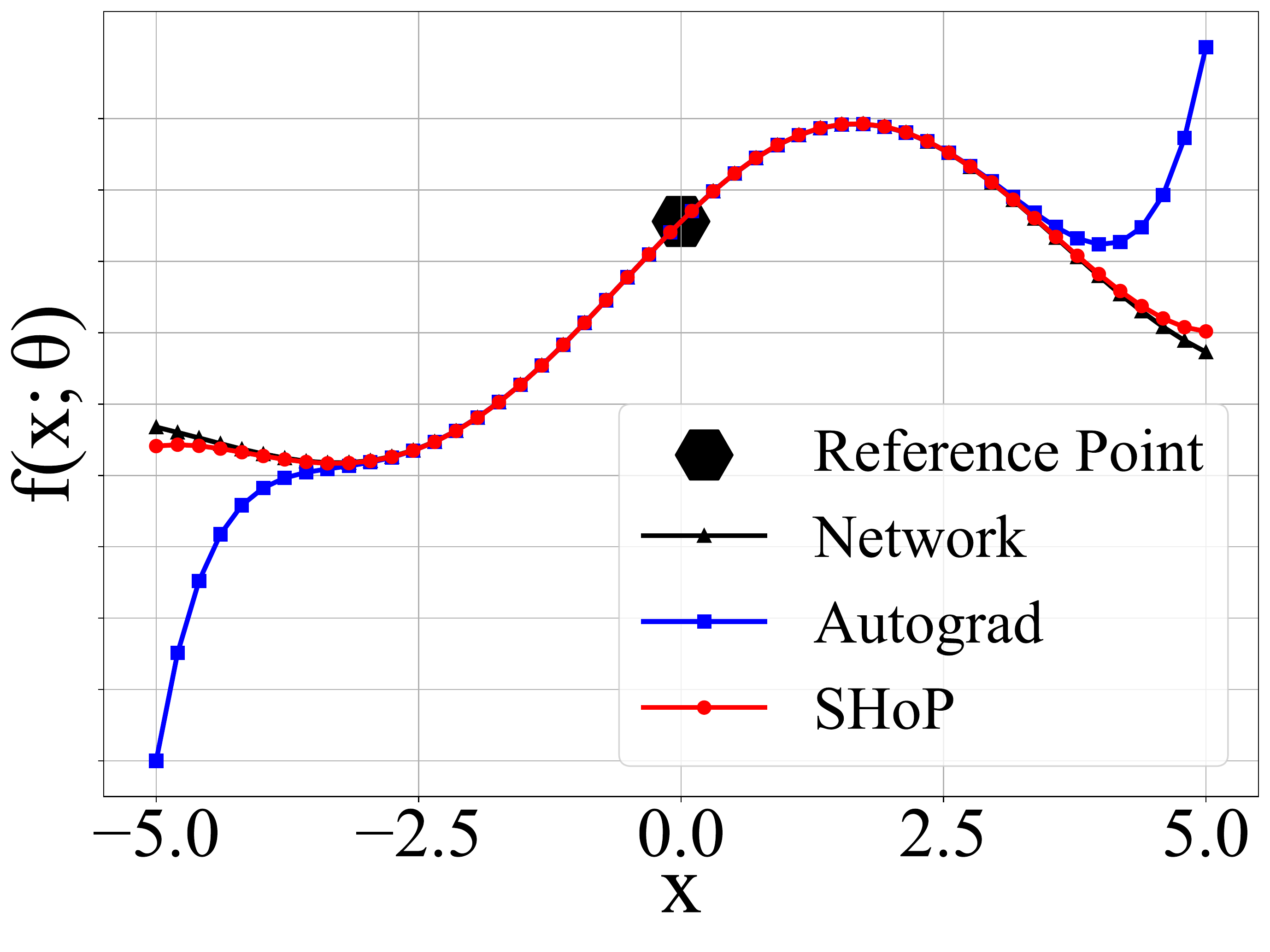}
    }
    \vspace{-3mm}
    \subfigure[]{
        \includegraphics[width=0.92\columnwidth]{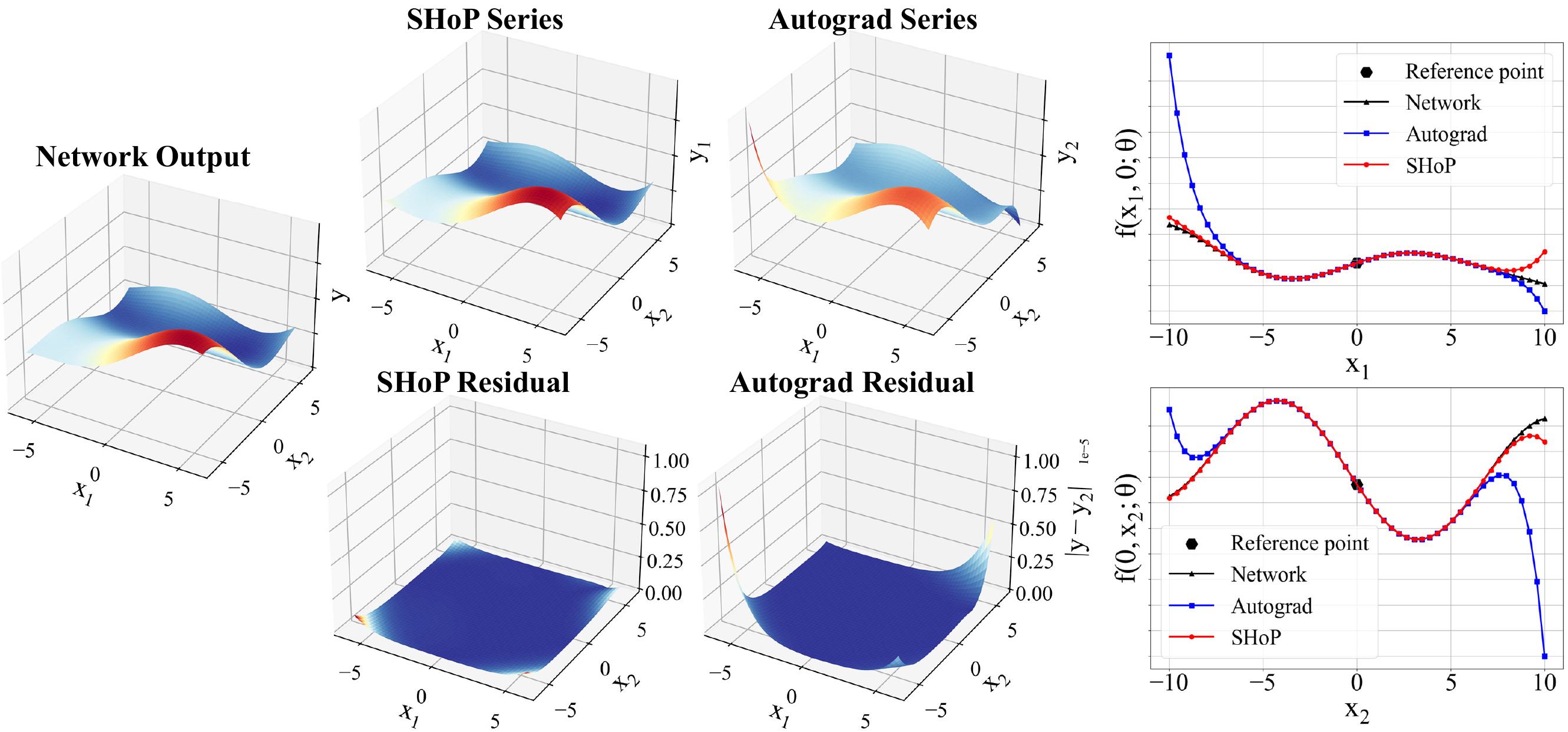}
    }
\caption{Approximation results of networks' Taylor expansion. (a) The approximation results of \M~ and Autograd on three different single-input MLPs.
(b) The performance on a 2D MLP. 
The leftmost panel shows the network output $y$. The four middle panels show the approximated surfaces (upper) and residuals (lower) of \M's polynomials $y_1$ and that of Autograd $y_2$, respectively. The two rightmost panels plot the profiles of the approximation accuracy at $x_2$=0 and $x_1$=0.} 
\label{expansion effect}
\end{center}
\vskip -6mm
\end{figure}

\section{Experiments}
\vspace{-2mm}
\subsection{Implementation details}\label{Implementation details}
\vspace{-1mm}
We use MLPs with Sine activation function\cite{sitzmann2020implicit} to validate our Taylor expansion of neural networks and its capability of solving high-order PDEs. Since the expansion of the multi-output model is a direct extension of the single-output model, here we use single-output setting for easier demonstration.  



We use Adamax\cite{kingma2014adam} to optimize the parameters and conduct 1000 epochs of model optimization in most cases. The learning rate is initialized to be 5e-3 and 
MultiStepLR is adopted to schedule the learning rate progressively. The \M~ framework is implemented with Pytorch, and the GPU version is NVIDIA GeForce RTX 3090 on a Linux operation system. For more implementation details, please refer to the Supplementary Materials. Our code will soon be publicly available at https://github.com/HarryPotterXTX/SHoP.

\vspace{-1mm}
\subsection{Performance of the high-order derivative rule}
\vspace{-2mm}
In this section, we test the accuracy and efficiency of the new high-order derivative rule, and compare it with the widely used Autograd. 

\noindent\textbf{Accuracy of derivatives.\quad}
Fig.~\ref{expansion effect}(a) shows the approximation results of \M~ and Autograd on three 1D MLPs. We use above two approaches to calculate the first 10-order derivatives at the reference points and use them to approximate the target MLPs. The plots show that our results (red curves) are closer to the true output (black curves), while Autograd (blue curves) fits well near the reference point but deviates a lot as the input moves far from the reference point. We can induce that although Autograd can calculate low-order derivatives well but is of insufficient accuracy when dealing with high-order derivatives. On the contrary, \M~ conducts one-step inference to avoid error accumulation and thus achieves high accuracy even at high-orders.
Fig.~\ref{expansion effect}(b) compares the approximated surfaces (middle, upper) and residues (middle, lower) of \M~ and Autograd on a 2D MLP with output shown in the left panel, and shows their results along $x_1$=0 and $x_2$=0 (right). Both plots display our superior performance and arrive at the same conclusion as in Fig. \ref{expansion effect}(a). 


\noindent\textbf{Running efficiency.\quad} 
Tab.~\ref{time table} shows the running time of our approach in parallel with that of Autograd for calculating the first $k$ order derivatives of a $p$-input MLP. With the increase of input dimension and order, the running time of both methods increase but our running time is consistently shorter than Autograd by a large margin. When $p$=2 and $k$=10, we just need 0.3828s, while Autograd takes 1435.0s. Besides, when $p$=3 and $k$=8, Autograd runs out of memory because 
it need to create too many computation graphs, while we finish it in just 0.2618s, which prove the higher time and memory efficiency of our method. The time complexity comparison of \M~ and Autograd can be found in Supplementary Materials.

\begin{table}[t]
\vspace{-4mm}
\caption{Comparison of the running time between \M~ and Autograd. Here $p$ denotes the dimension of input, and $k$ is the order of derivatives, and ``OOM'' means out of memory.}\label{time table}
\vspace{-4mm}
\begin{center}
\resizebox{\columnwidth}{!}{
    \begin{scriptsize}
    \setlength{\tabcolsep}{0.5mm}{
        \begin{tabular}{cc||cccccccccc}
        \hline
        \multicolumn{2}{c||}{\quad\quad\quad$k$\quad\quad\quad} & \quad\quad1\quad\quad & \quad\quad2\quad\quad & \quad\quad3\quad\quad & \quad\quad4\quad\quad & \quad\quad5\quad\quad & \quad\quad6\quad\quad & \quad\quad7\quad\quad & \quad\quad8\quad\quad & \quad\quad9\quad\quad & \quad\quad10\quad\quad \\ 
        \hline
        \multicolumn{1}{c|}{\multirow{2}{*}{$p$=1}} & \M     
        & 0.0252 & 0.0303 & 0.0397 & 0.0526 & 0.0619 & 0.0935 & 0.1343 & 0.1801 & 0.2319 & 0.3480 \\
        \multicolumn{1}{c|}{} & Autograd 
        & 0.0437 & 0.0450 & 0.0474 & 0.0556 & 0.0706 & 0.1092 & 0.2151 & 0.4524 & 1.3157 & 4.1036 \\ 
        \hline
        \multicolumn{1}{c|}{\multirow{2}{*}{$p$=2}} & \M
        & 0.0267 & 0.0321 & 0.0408 & 0.0538 & 0.0715 & 0.1043 & 0.1356 & 0.1922 & 0.2982 & 0.3828 \\
        \multicolumn{1}{c|}{} & Autograd 
        & 0.0472 & 0.0485 & 0.0532 & 0.0756 & 0.1958 & 0.8930 & 5.1107 & 32.028 & 204.34 & 1435.0 \\ 
        \hline
        \multicolumn{1}{c|}{\multirow{2}{*}{$p$=3}} & \M
        & 0.0272 & 0.0322 & 0.0418 & 0.0567 & 0.0720 & 0.1080 & 0.1623 & 0.2618 & 0.5235 & 1.2327 \\
        \multicolumn{1}{c|}{} & Autograd 
        & 0.0457 & 0.0514 & 0.1102 & 0.4226 & 2.6690 & 24.417 & 236.75 & OOM & OOM & OOM \\ 
        \hline
        \end{tabular}
    }
    \end{scriptsize}
    }
\end{center}
\vskip -0.1in
\end{table}

\noindent\textbf{Convergence under different parameter settings.\quad} In Tab.~\ref{converge table}, we initialized the weights of each layer following uniform distribution $\mathbf{W}_{i,j}^{(m)}\sim U(-\frac{w_0}{o_{m-1}},\frac{w_0}{o_{m-1}})$. When $w_0=0.01,~0.1$, the higher-order derivatives are far smaller than the lower-order derivatives, and we can ignore the higher-order derivatives and the Taylor series converges. When $w_0=1.0$, the derivatives of different orders oscillate, and the higher-order terms cannot be ignored. When $w_0=10,~100$, Taylor series are seriously divergent. 
The results inspire us to impose proper constraints on the network parameters when using its Taylor series as a surrogate for either calculation or analysis. 
We can also induce that the neural networks' strong capability of fitting diverse functions is attributed to its wide Taylor series covering all convergence cases. 

\begin{table}[t]
\caption{Convergence of the Taylor series of a neural network (depth 3, width 32) under different parameter distributions. Here $w_0$ determines the distribution of network parameters $U(-\frac{w_0}{o_{m-1}},\frac{w_0}{o_{m-1}})$, and the scores in each cell is 
$|\frac{\partial^k f}{\partial x^k}/\frac{\partial f}{\partial x}|$, indicating the convergence.}
\label{converge table}
\begin{center}
\vspace{-2mm}
\resizebox{\columnwidth}{!}{
    \begin{scriptsize}
    \setlength{\tabcolsep}{1.0mm}{
    \begin{tabular}{c||cccccccccc} 
        \hline
            $w_0$ & $n$=1 & $n$=2 & $n$=3 & $n$=4 & $n$=5 & $n$=6 & $n$=7 & $n$=8 & $n$=9 & $n$=10\\ 
        \hline
            0.010 & 1.00e+00 & 6.16e-03 & 4.77e-05 & 3.80e-07 & 2.51e-09 & 2.57e-11 & 1.44e-13 & 1.81e-15 & 8.79e-18 & 1.32e-19 \\
            0.100 & 1.00e+00 & 3.74e-02 & 6.87e-03 & 1.88e-04 & 5.60e-05 & 1.93e-06 & 4.79e-07 & 1.97e-08 & 4.22e-09 & 1.91e-10 \\
            1.000 & 1.00e+00 & 0.67e+00 & 0.36e+00 & 0.37e+00 & 0.28e+00 & 0.25e+00 & 0.48e+00 & 0.06e+00 & 1.22e+00 & 0.62e+00 \\
            10.00 & 1.00e+00 & 4.58e+01 & 1.02e+02 & 7.31e+03 & 3.68e+04 & 1.72e+06 & 1.69e+07 & 5.20e+08 & 1.05e+10 & 1.96e+11 \\
            100.0 & 1.00e+00 & 6.29e+03 & 3.90e+05 & 7.86e+08 & 2.54e+11 & 2.97e+14 & 2.93e+17 & 1.57e+20 & 4.96e+23 & 2.25e+26 \\
    \bottomrule
    \end{tabular}
    }
    \end{scriptsize}
    }
\end{center}
\vskip -0.2in
\end{table}

\vspace{-1mm}
\subsection{Effectiveness of solving PDEs}
\vspace{-2mm}
Here we demonstrate the performance of our method on four different types of PDEs: a 1D 4th-order Harmonic oscillator system, a 2D 4th-order Biharmonic equation, a 2D 8th-order Helmholtz equation, and a 3D 4th-order Heat equation. The detailed PDE conditions, initial conditions, and boundary conditions can be found in Supplementary Materials.

\begin{figure}[t]
\vspace{-4mm}
\begin{center}
\subfigure[]{
        \includegraphics[width=0.24\columnwidth]{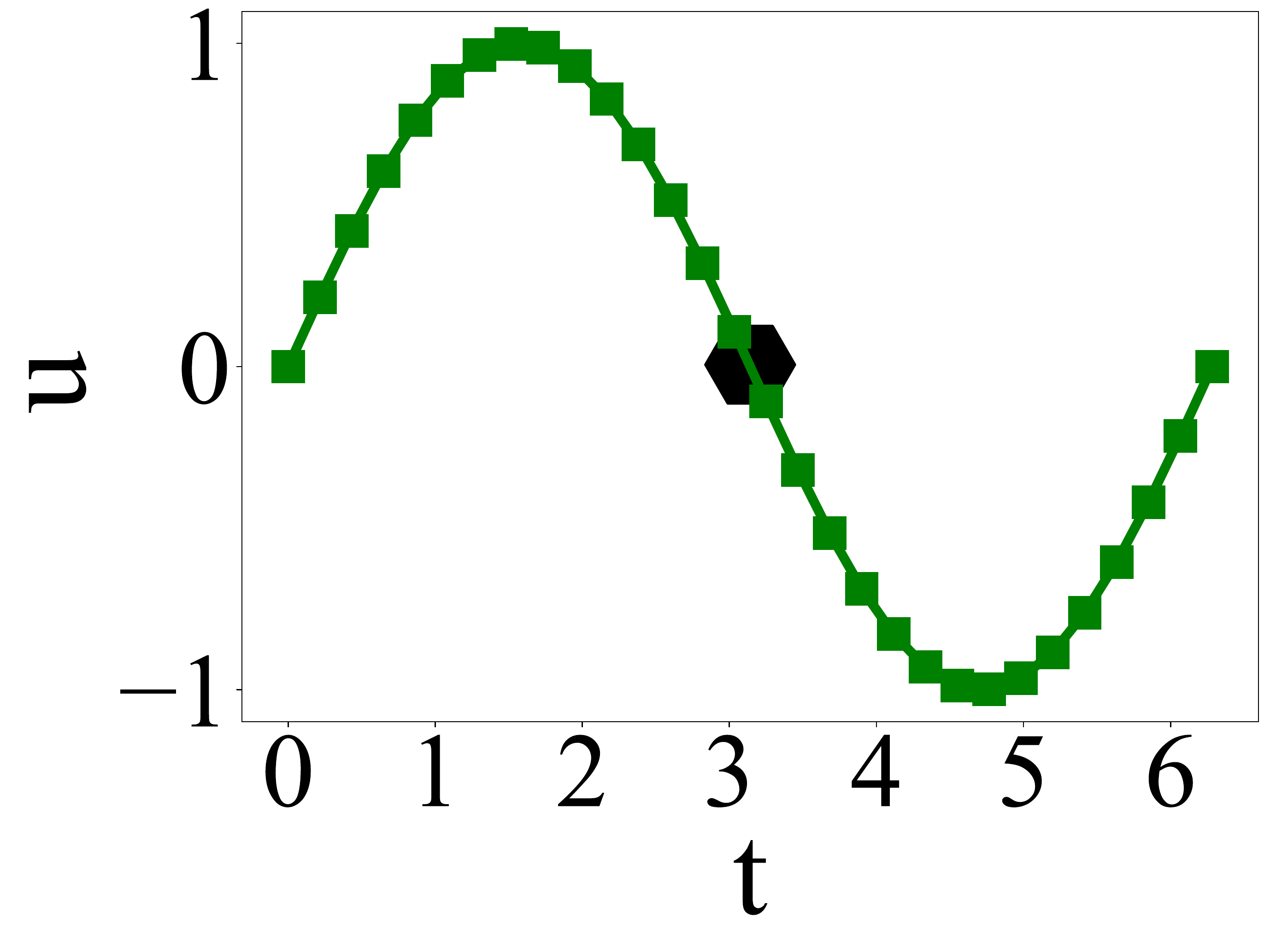}
    }
    \hspace{-0.1in}
\subfigure[]{
        \includegraphics[width=0.24\columnwidth]{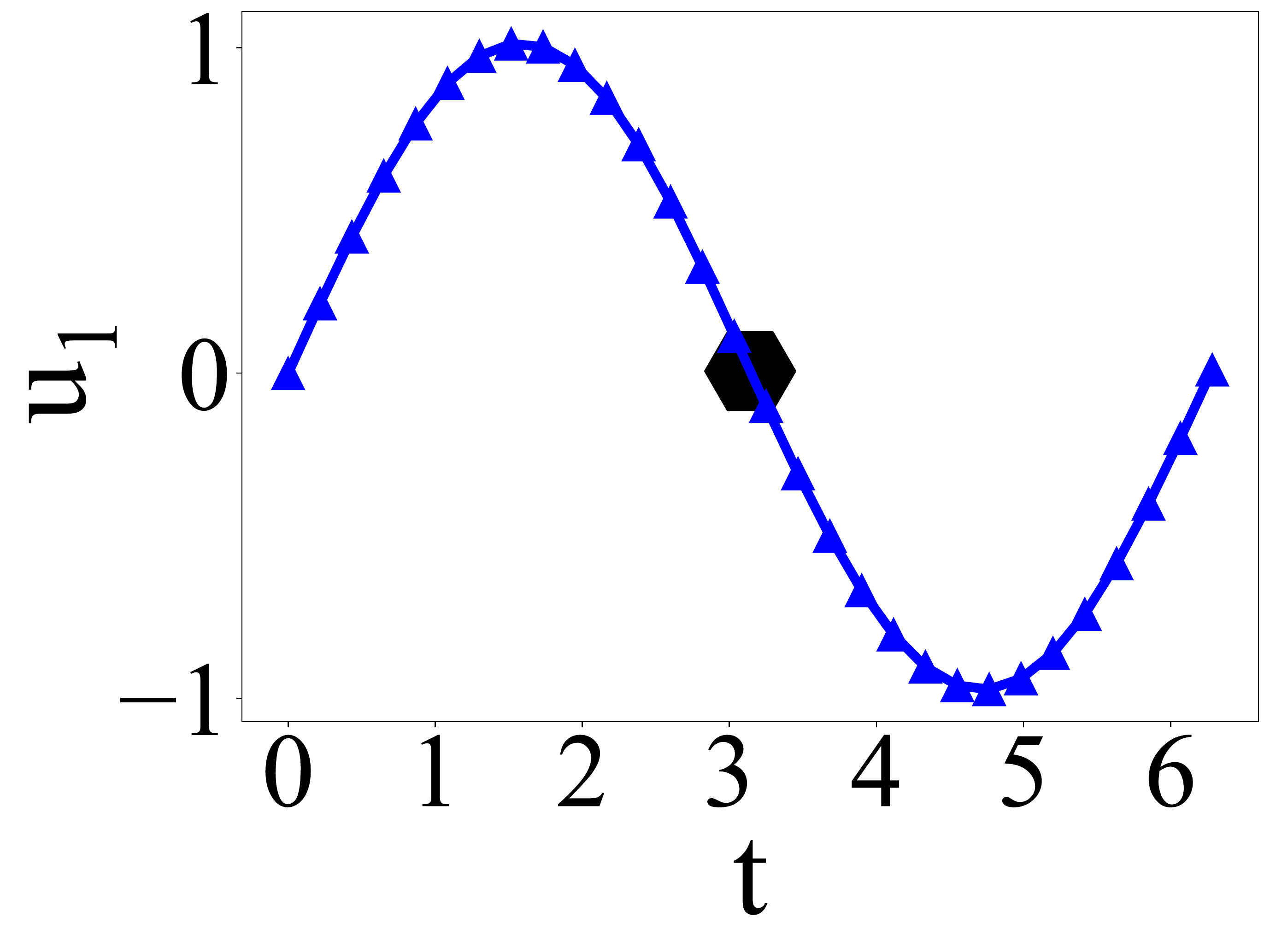}
    }
    \hspace{-0.1in}
\subfigure[]{
        \includegraphics[width=0.24\columnwidth]{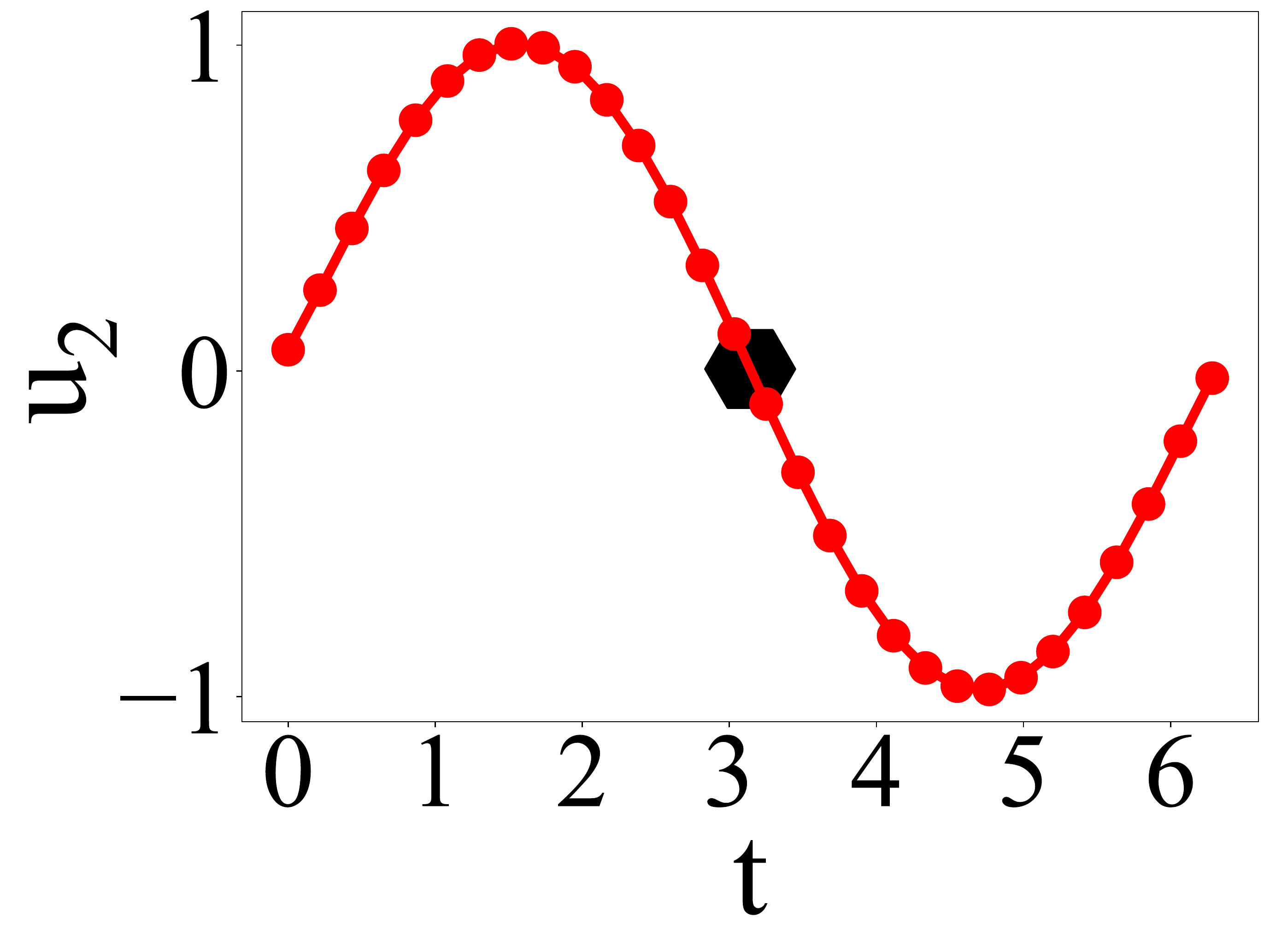}
    }
    \hspace{-0.1in}
\subfigure[]{
        \includegraphics[width=0.24\columnwidth]{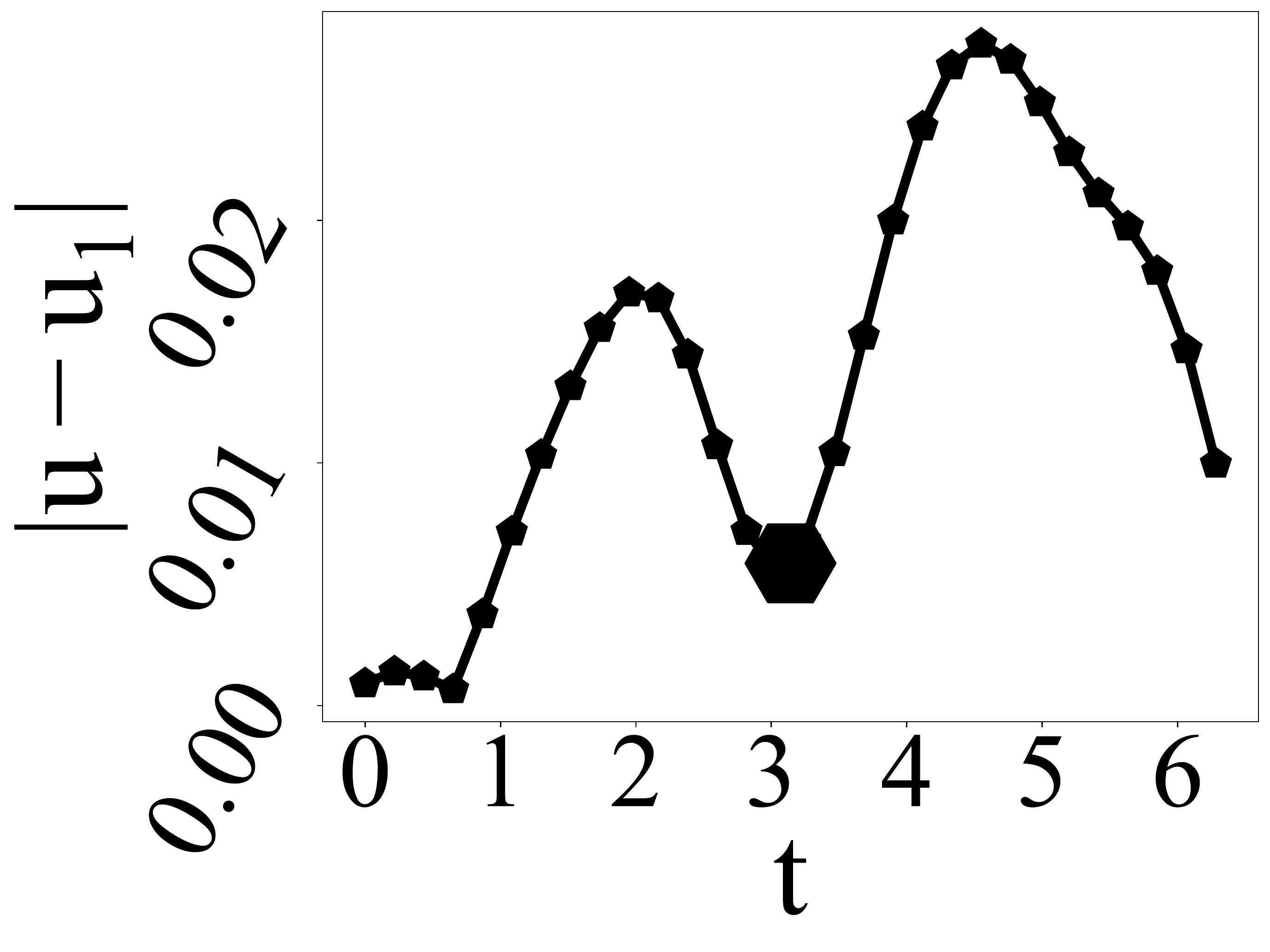}
    }    
\vspace{-3mm}
\caption{\textbf{Performance on 1D 4th-order Harmonic oscillator system.} (a) The true solution $u$ in interval [0, 2$\pi$]. (b) \M's solution $u_1$. (c) The Taylor expansion result $u_2$ for the network on the reference input $t=\pi$. (d) The approximation error $|u-u_1|$.}
\label{sin}
\end{center}
\vskip -0.2in
\end{figure}

\noindent\textbf{1D function.\quad} We consider a 1D 4th-order Harmonic oscillator system
\begin{equation}  
\begin{split}
\left\{  
    \begin{array}{ll}
    u_{tttt}+2u_{tt}+u=0, & t\in [0,2\pi], \\
    u(0)=0, u_t(0)=1, u_{tt}(0)=0.
    \end{array}  
\right.  
\end{split}
\end{equation} 
The initial conditions indicate that the initial position of the harmonic oscillator is the balance point, the initial speed is 1, and the initial acceleration is 0. The hyper-parameters $\lambda=5$, $\mu=1$, and the results are shown in Fig.~\ref{sin}.

The results show that the solution obtained by \M~($u_1(t)$, blue) is very close to the true solution ($u(t)$, green), and we expand the final black-box network into a 10-order Taylor polynomial ($u_2(t)$, red), providing an explicit explanation for the final solution. 
From the output of the network on input $\pi$ and the first 10 order derivatives, the Taylor polynomial can be described as
\begin{equation}
\begin{split}
    u_2(t)&=0.0059-\f{0.9943}{1!}\Delta_{\pi}+\f{0.0097}{2!}\Delta_{\pi}^2+\f{0.9986}{3!}\Delta_{\pi}^3-\f{0.0235}{4!}\Delta_{\pi}^4-\f{1.0089}{5!}\Delta_{\pi}^5\\
    &+\f{0.0680}{6!}\Delta_{\pi}^6+\f{1.0894}{7!}\Delta_{\pi}^7-\f{0.2868}{8!}\Delta_{\pi}^8-\f{1.9500}{9!}\Delta_{\pi}^9+\f{1.5491}{10!}\Delta_{\pi}^{10},
\end{split}
\end{equation}
where $\Delta_{\pi}=t-\pi$. One can easily see that the network outputs match well with the Taylor coefficients of the true solution $u=sin(t)$, which again validates the accuracy of \M's solution.

\begin{figure}[t]
\vspace{-4mm}
\begin{center}
\subfigure[]{
        \includegraphics[width=0.24\columnwidth]{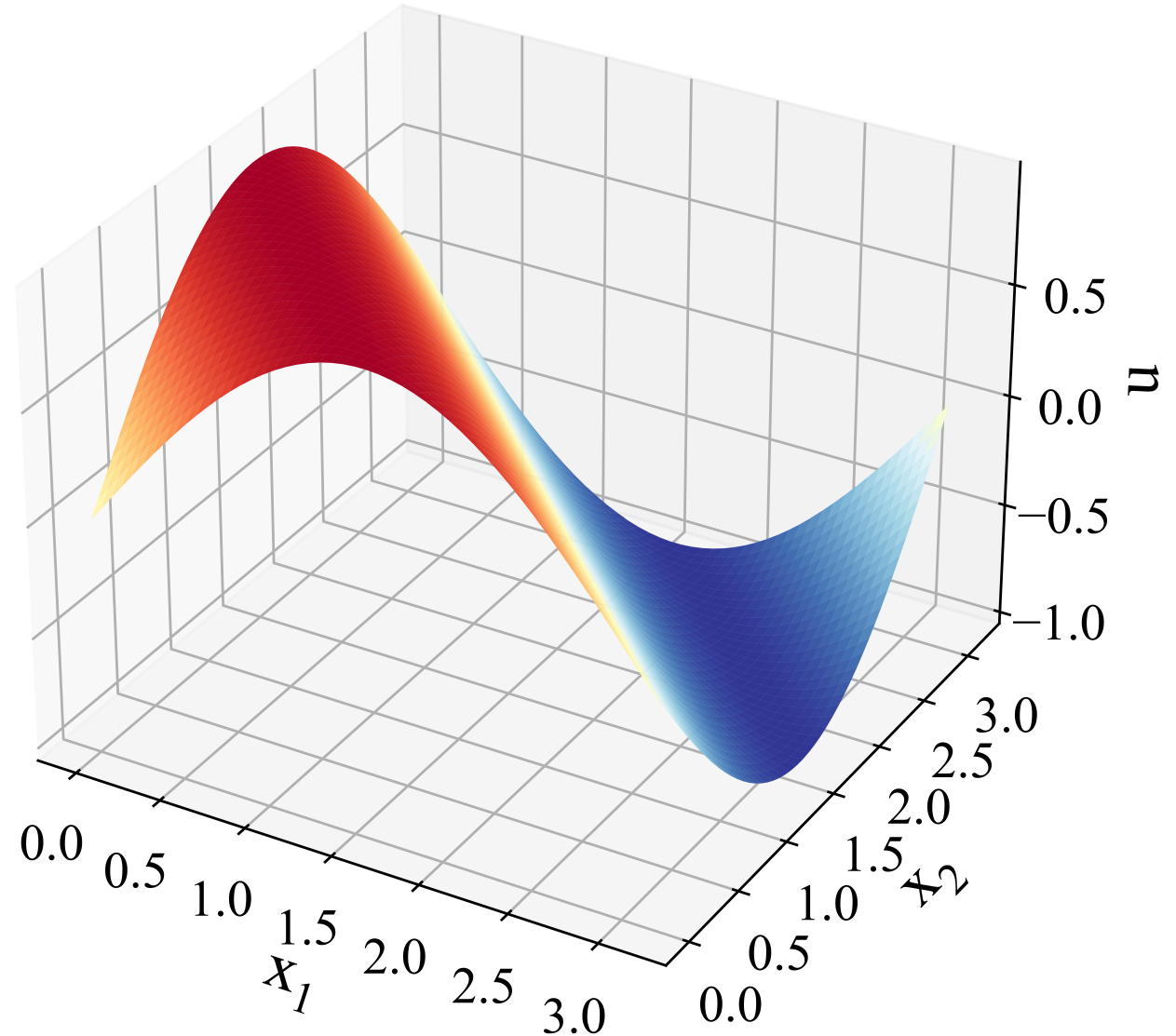}
    }
    \hspace{-0.1in}
\subfigure[]{
        \includegraphics[width=0.24\columnwidth]{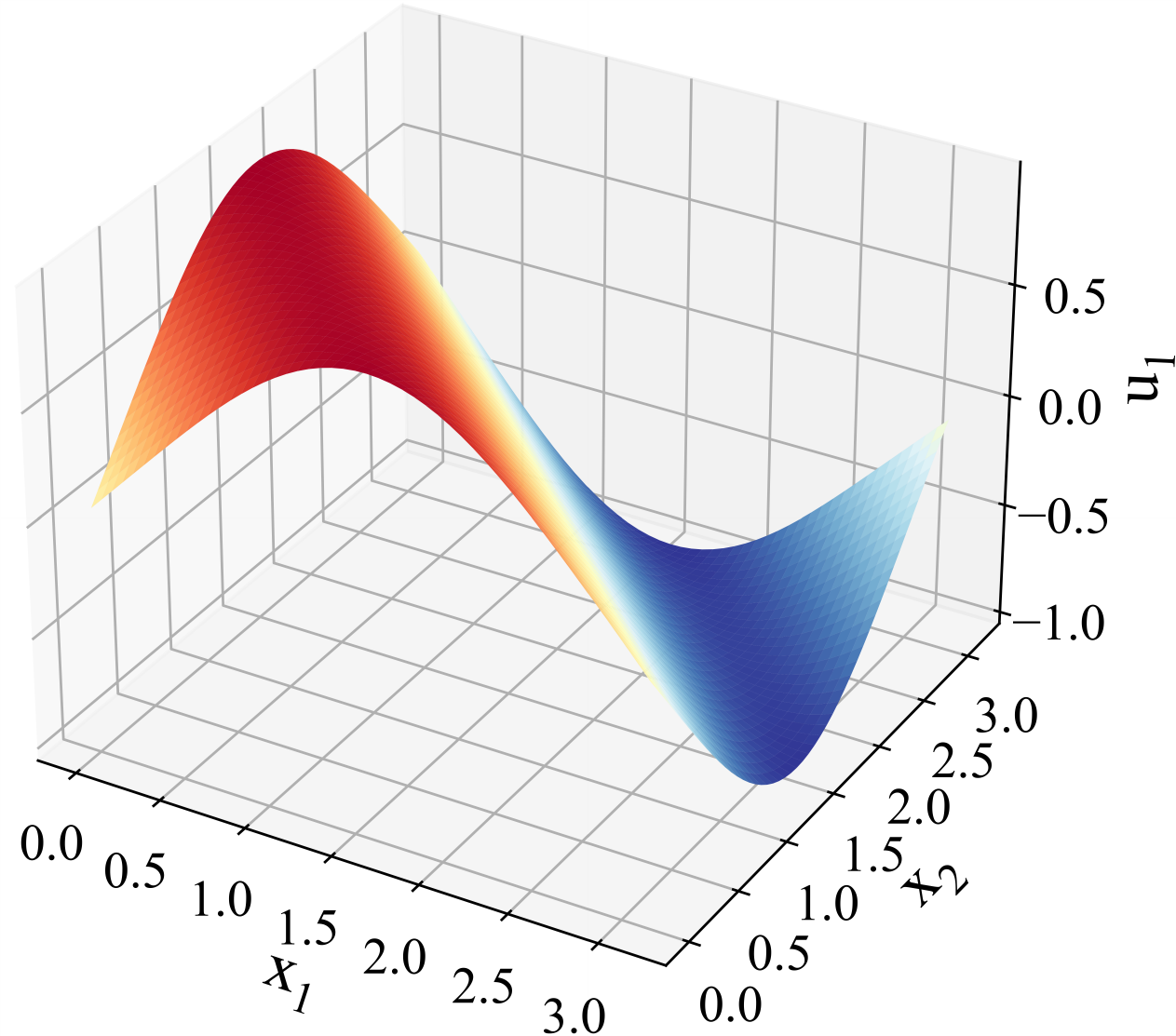}
    }
    \hspace{-0.1in}
\subfigure[]{
        \includegraphics[width=0.24\columnwidth]{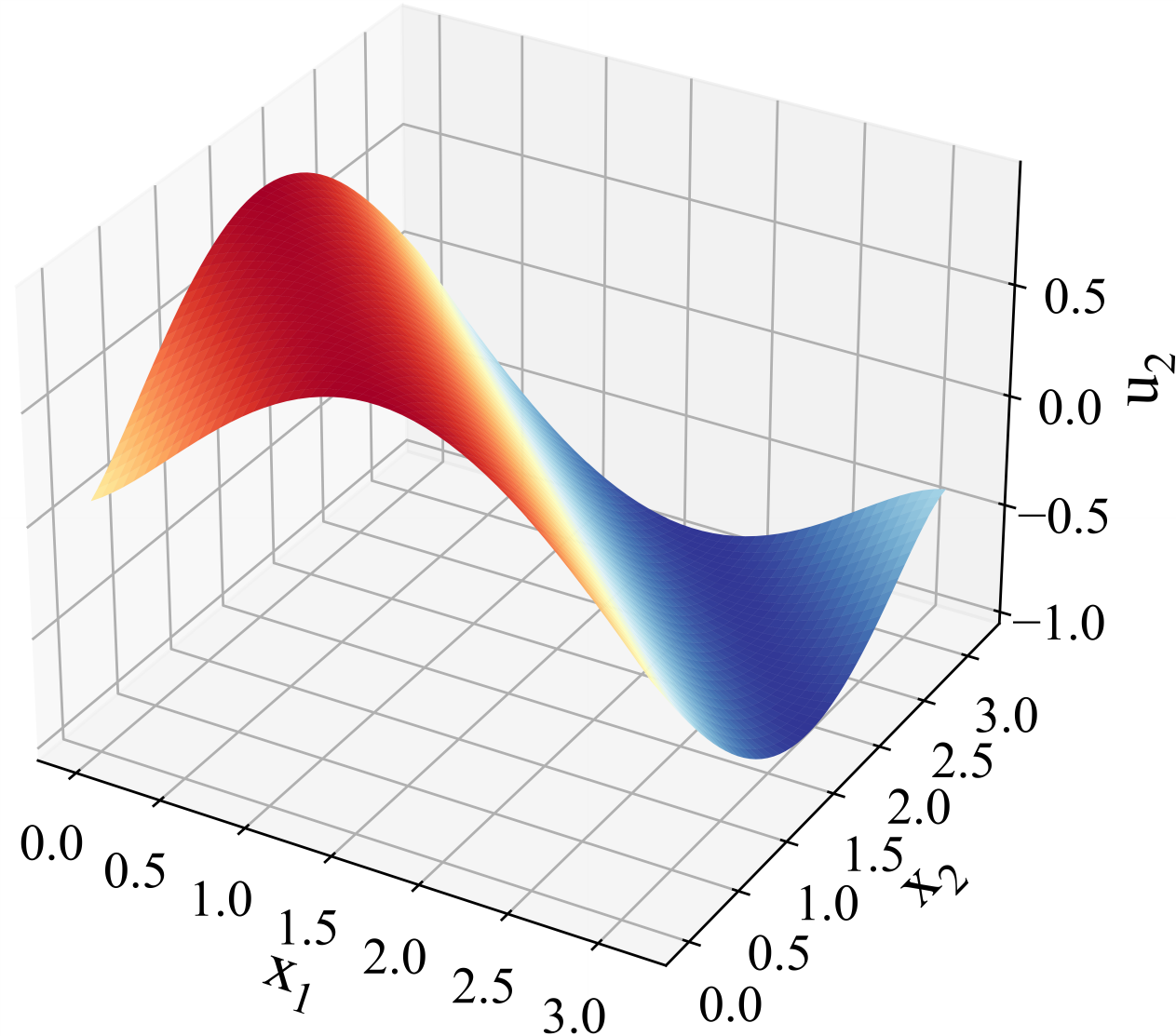}
    }
    \hspace{-0.1in}
\subfigure[]{
        \includegraphics[width=0.24\columnwidth]{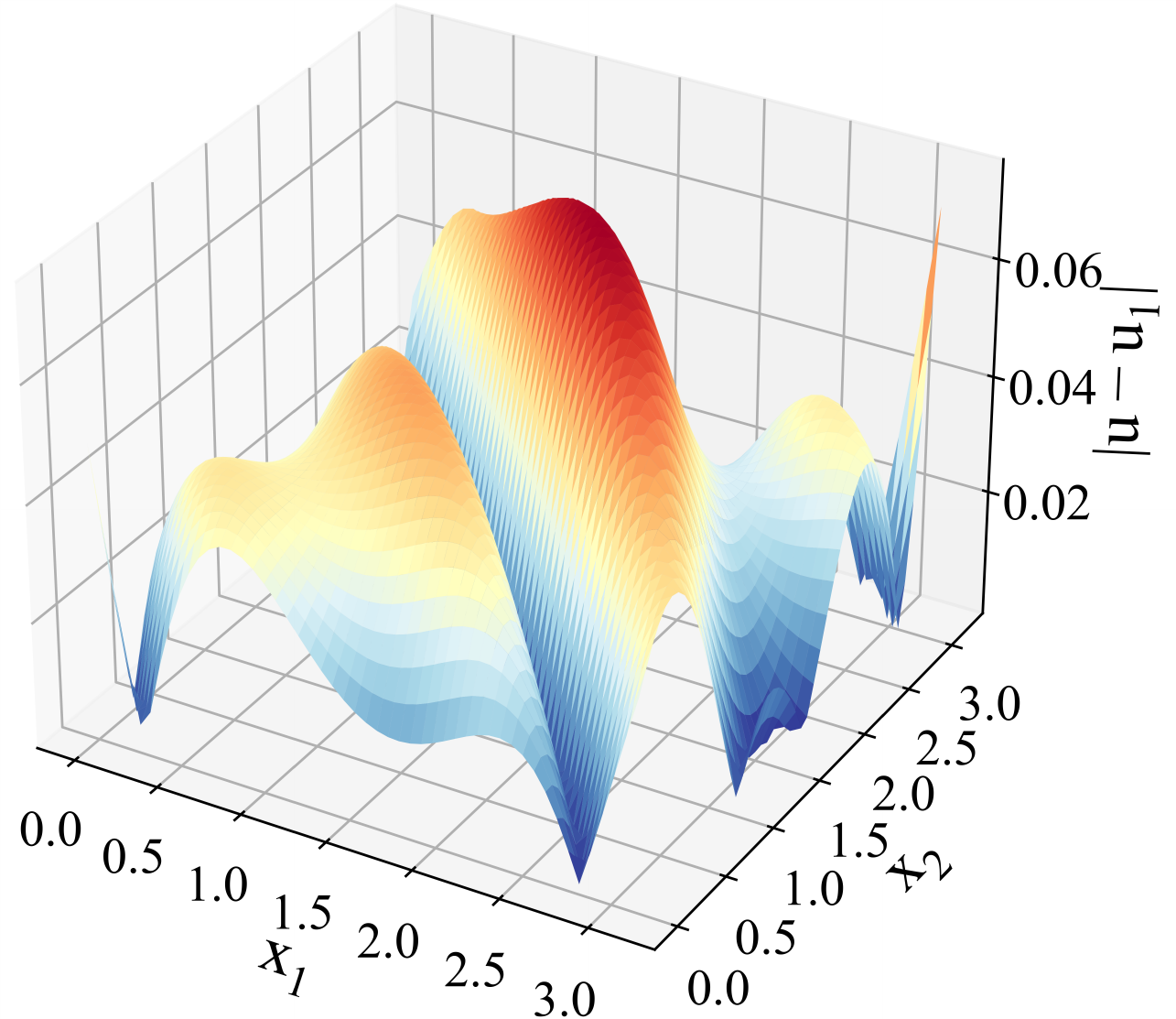}
    }    
\vspace{-2mm}
\caption{\textbf{Performance on 2D 4th-order Biharmonic equation.} (a) The true solution $u$ in interval $[0,\pi]\times [0,\pi]$. (b) \M's solution $u_1$. (c) The network's Taylor expansion result $u_2$ on the reference input $(x_1,x_2)=(0.5\pi,0.5\pi)$. (d) The approximation error  $|u-u_1|$.}
\label{2D 4order}
\end{center}
\vskip -0.2in
\end{figure}

\vspace{0.5mm}
\noindent\textbf{2D function.\quad} We solve a 2D fourth-order PDE and a 2D eighth-order PDE using \M.
The first one is a Biharmonic equation defined over $(x_1,x_2)\in [0,\pi]^2$, with PDE condition 
\begin{equation}  
    \nabla^4u=4sin(x_1+x_2),
\end{equation} 
where $\nabla^4$ is the fourth power of the del operator and the square of the Laplacian operator $\nabla ^{2}$ (or $\Delta$). 
Fig.~\ref{2D 4order} shows \M's performance on this PDE, with its solution close to the ground truth version. We also expand the network into a 2D 10-order Taylor polynomial on an inference input $(x_1,x_2)=(0.5\pi,~0.5\pi)$, and plot its output in Fig.~\ref{2D 4order}(c). The plot shows that the Taylor polynomial can actually provide a good approximation expression explicitly, making this neural network more transparent and interpretable.

The second one is a Helmholtz equation defined over $(x_1,x_2)\in [0,1]^2$, and its PDE condition is
\begin{equation}  
    \Delta^4u+u=17e^{-x_1-x_2}. 
\end{equation} 

\begin{figure}[t]
\begin{center}
\subfigure[]{        \includegraphics[width=0.24\columnwidth]{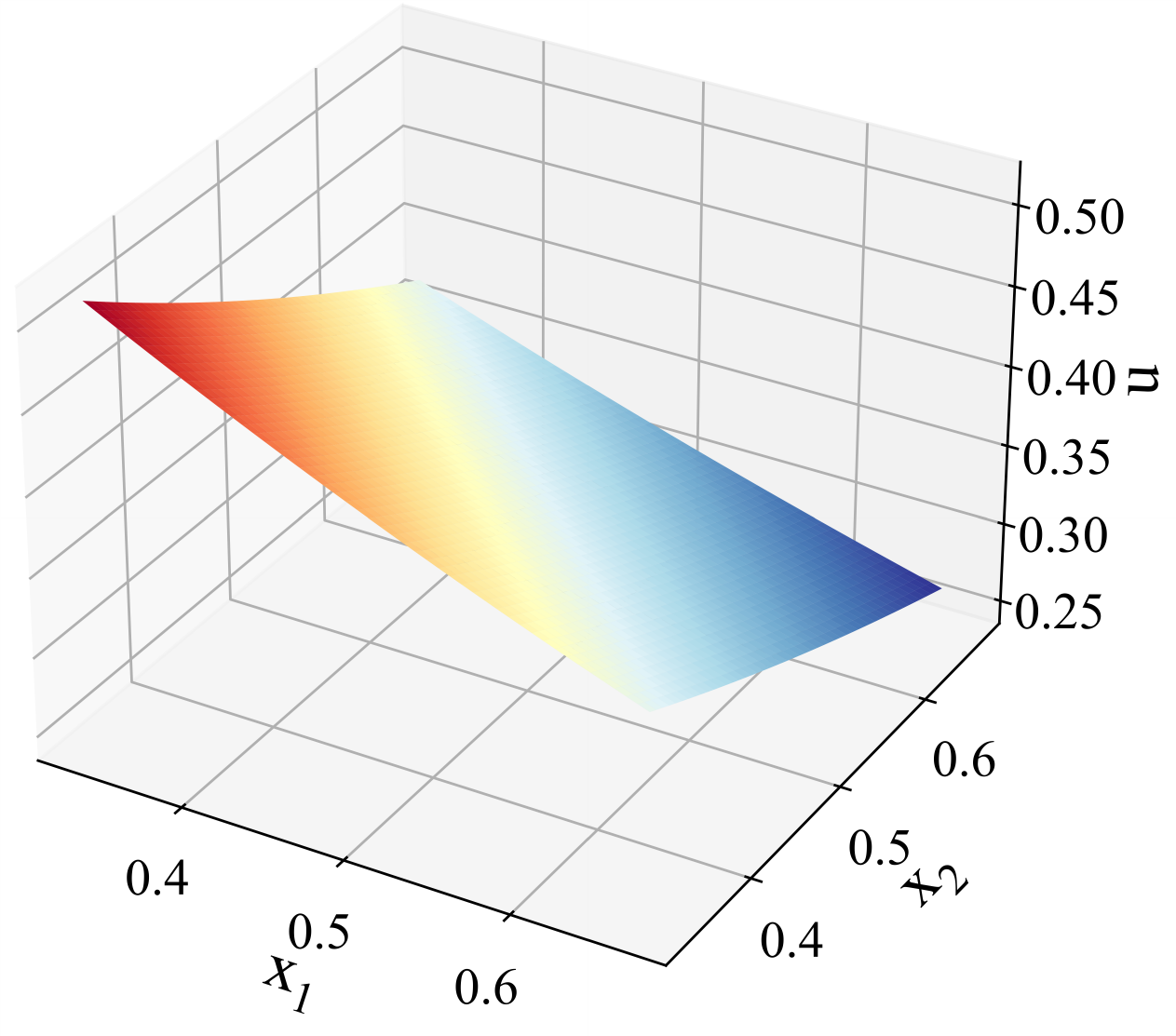}
    }
    \hspace{-0.1in}
\subfigure[]{        \includegraphics[width=0.24\columnwidth]{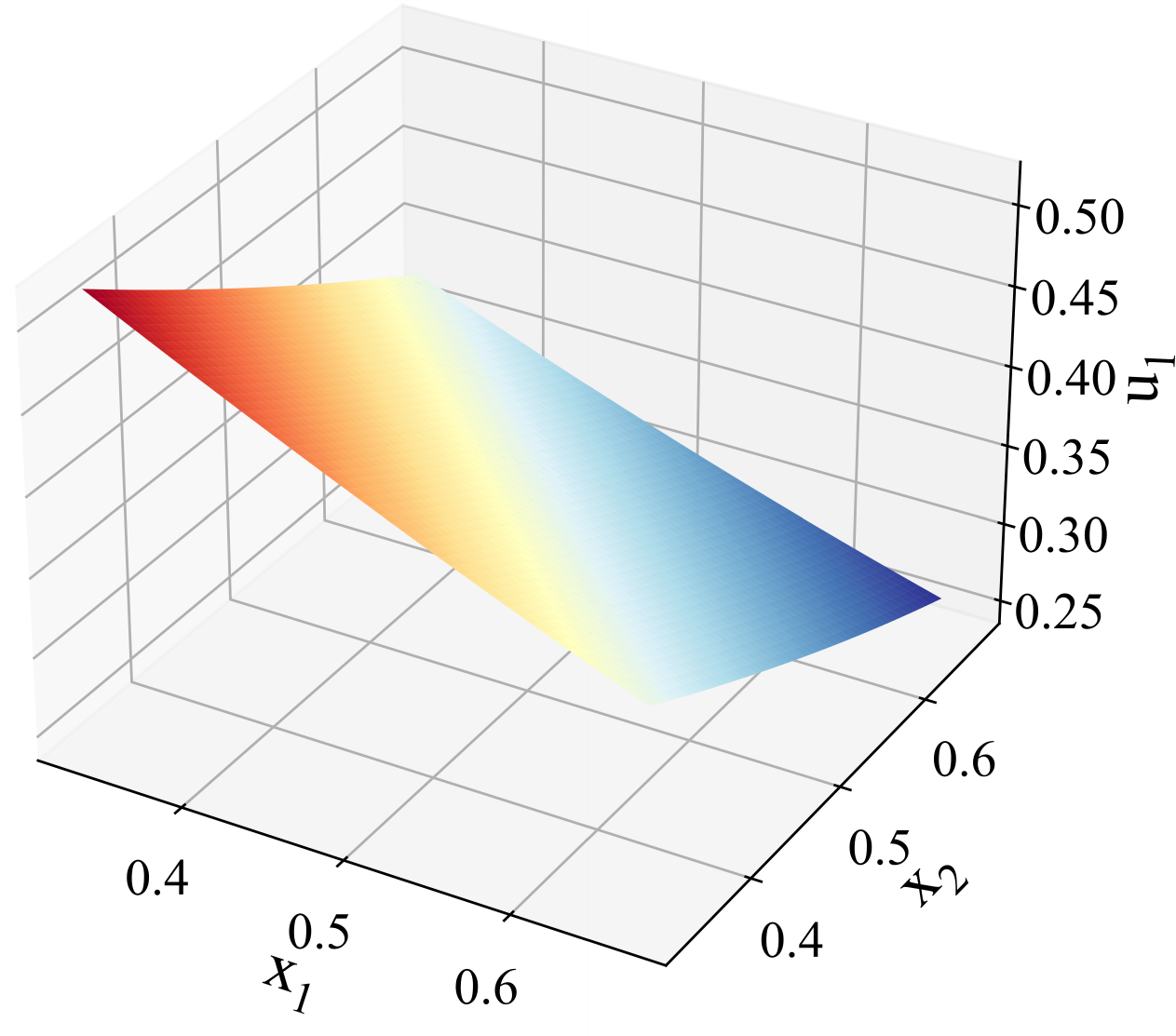}
    }
    \hspace{-0.1in}
\subfigure[]{
        \includegraphics[width=0.24\columnwidth]{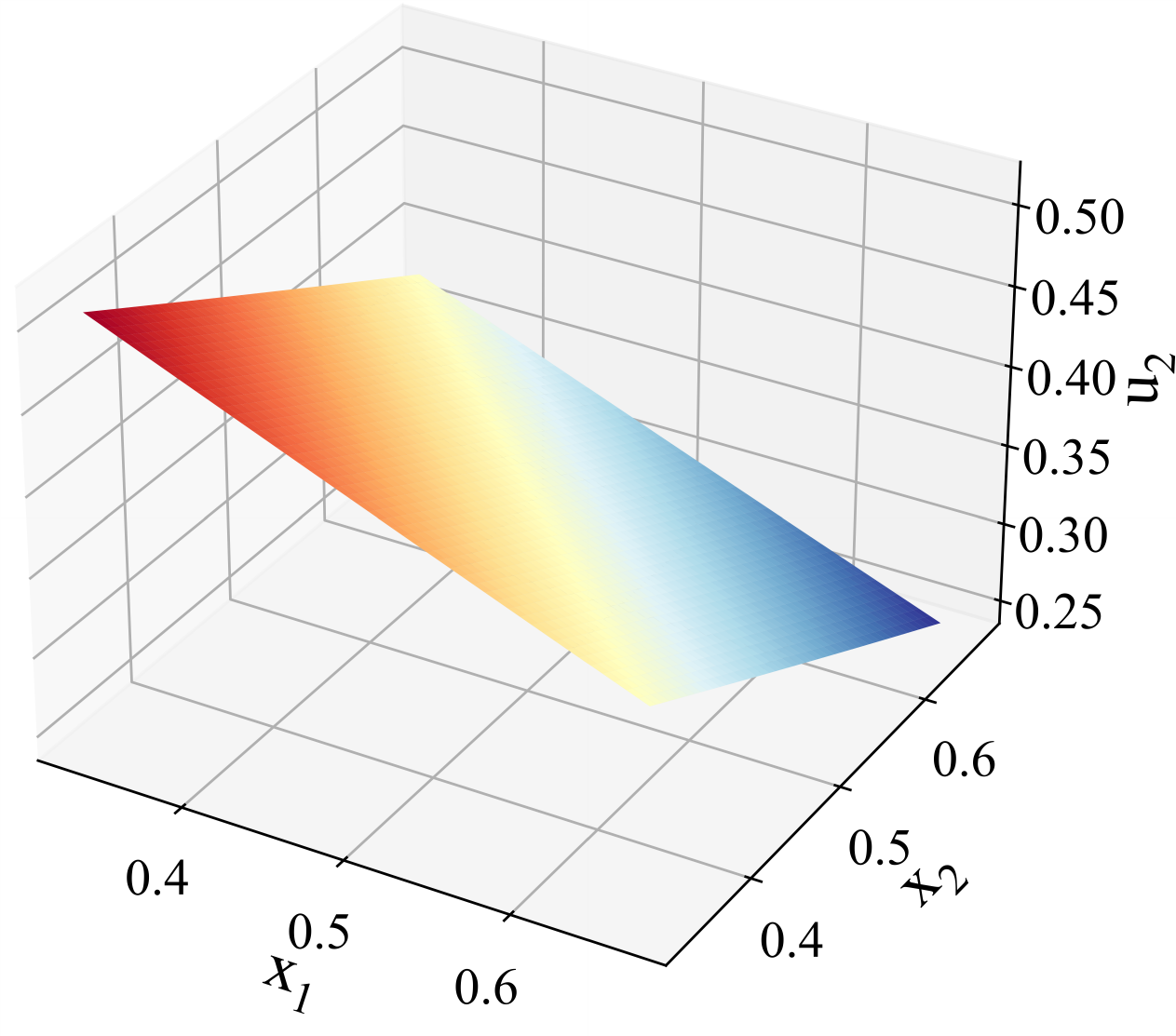}
    }
    \hspace{-0.1in}
\subfigure[]{
        \includegraphics[width=0.24\columnwidth]{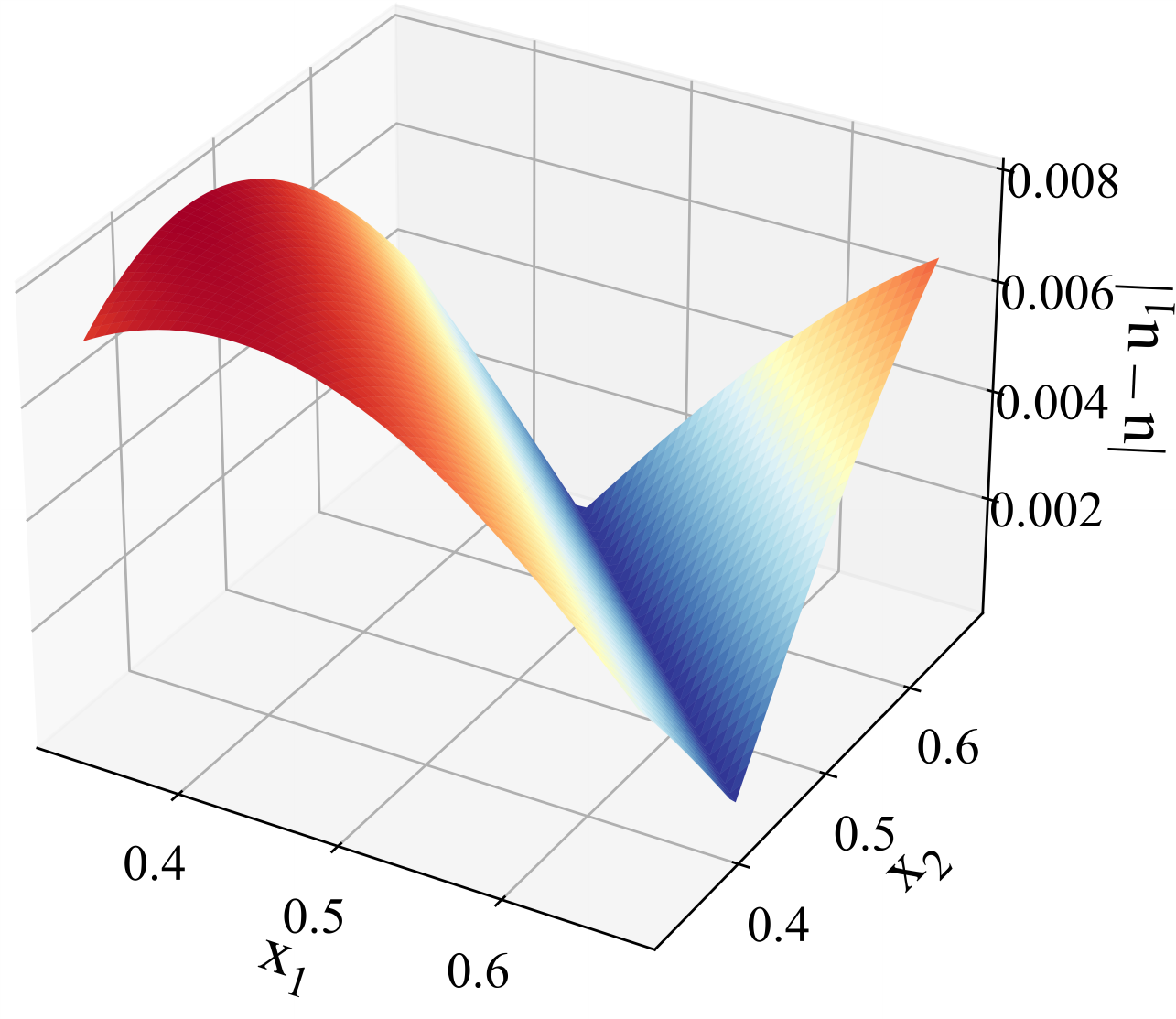}
    }    
\vspace{-2mm}
\caption{\textbf{Performance on 2D 8th-order Helmholtz equation.} (a) The true solution $u$ in interval $[\f{1}{3},\f{2}{3}]\times [\f{1}{3},\f{2}{3}]$. (b) \M's solution $u_1$. (c) The Taylor expansion result $u_2$ for the network on the reference input $(x_1,x_2)=(0.5,0.5)$. (d) The approximation error  $|u-u_1|$.}
\label{2D 8order}
\end{center}
\vskip -0.2in
\end{figure}

The results are shown in Fig.~\ref{2D 8order}. In line with the conclusion for the Biharmonic equation, \M~ has been shown to be an effective tool for obtaining approximate solutions to PDEs, and expanding the network into a polynomial help to derive an explicit solution for the PDE.

\begin{figure}[H]
\begin{center}
\subfigure[]{
        \includegraphics[width=0.23\columnwidth]{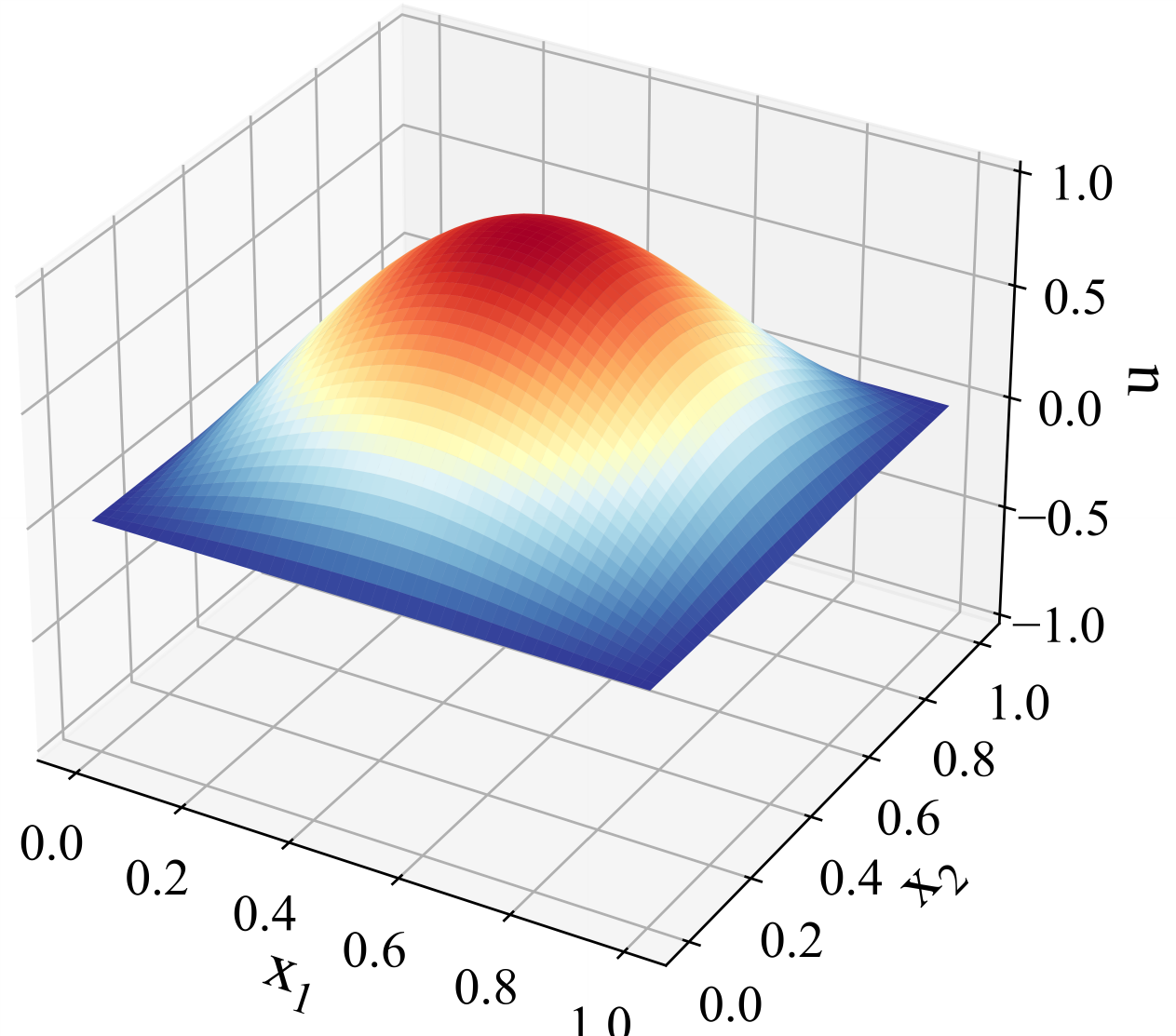}
    }
    \hspace{-0.1in}
\subfigure[]{
        \includegraphics[width=0.23\columnwidth]{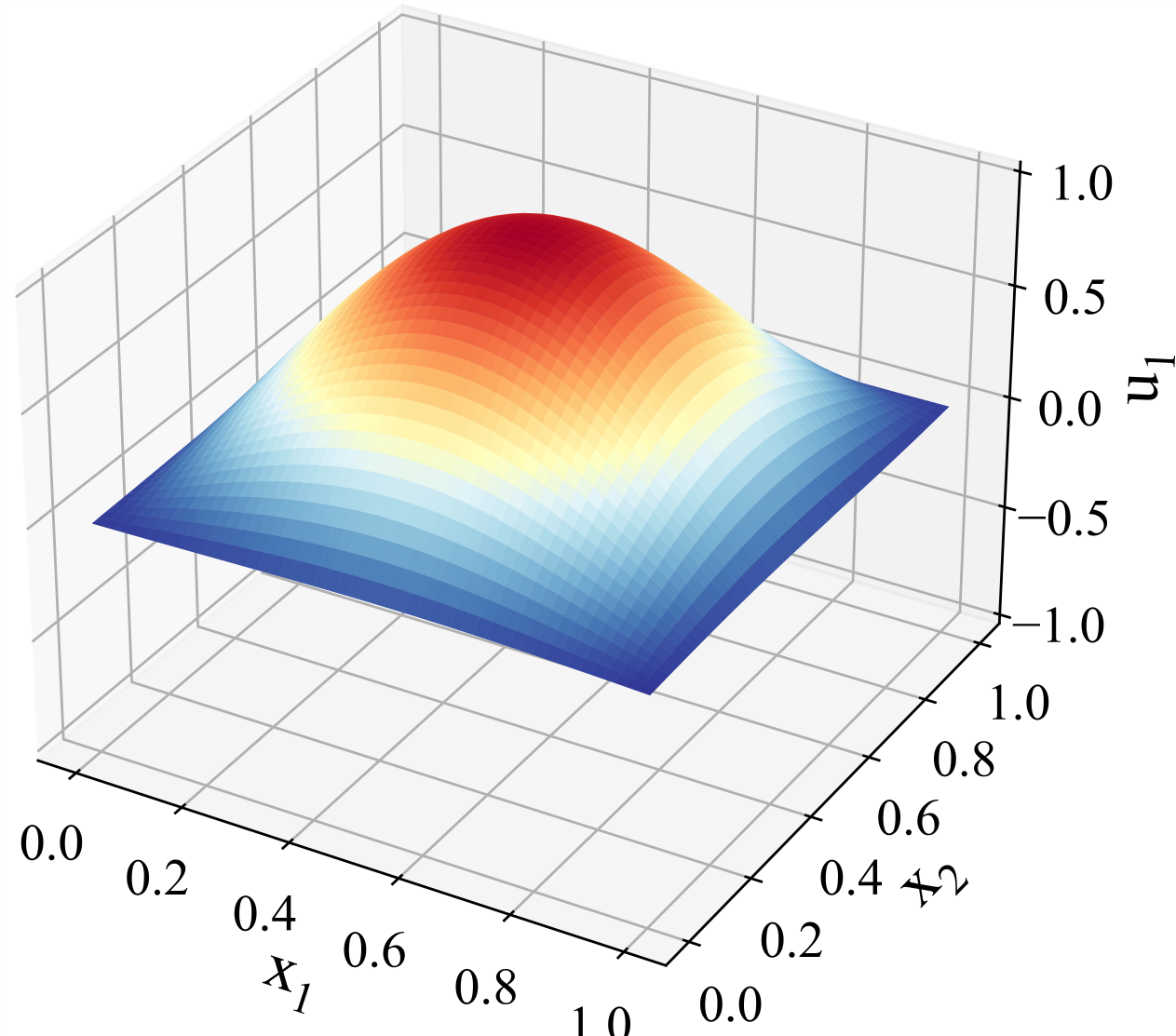}
    }
    \hspace{-0.1in}
\subfigure[]{
        \includegraphics[width=0.23\columnwidth]{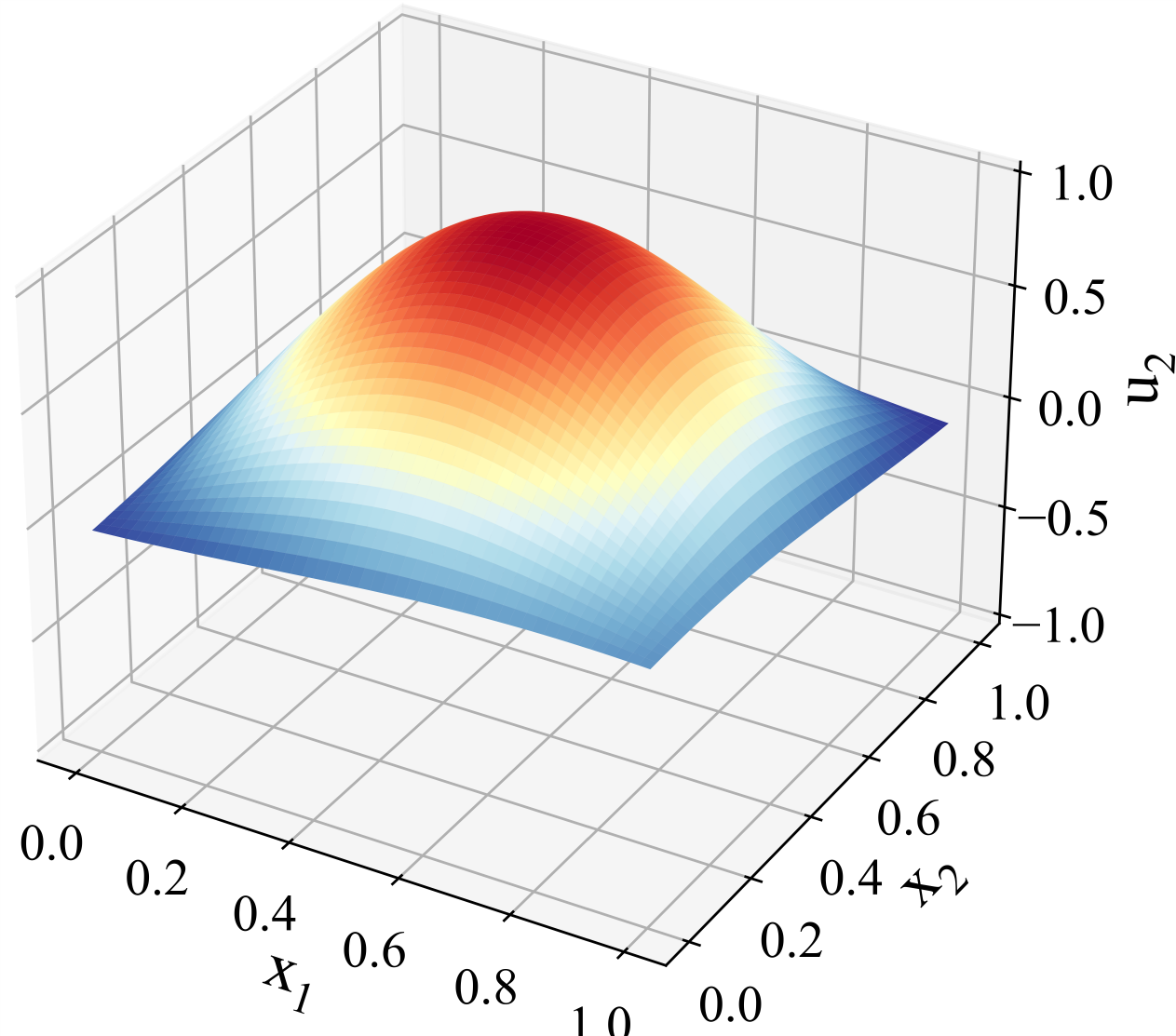}
    }
    \hspace{-0.1in}
\subfigure[]{
        \includegraphics[width=0.23\columnwidth]{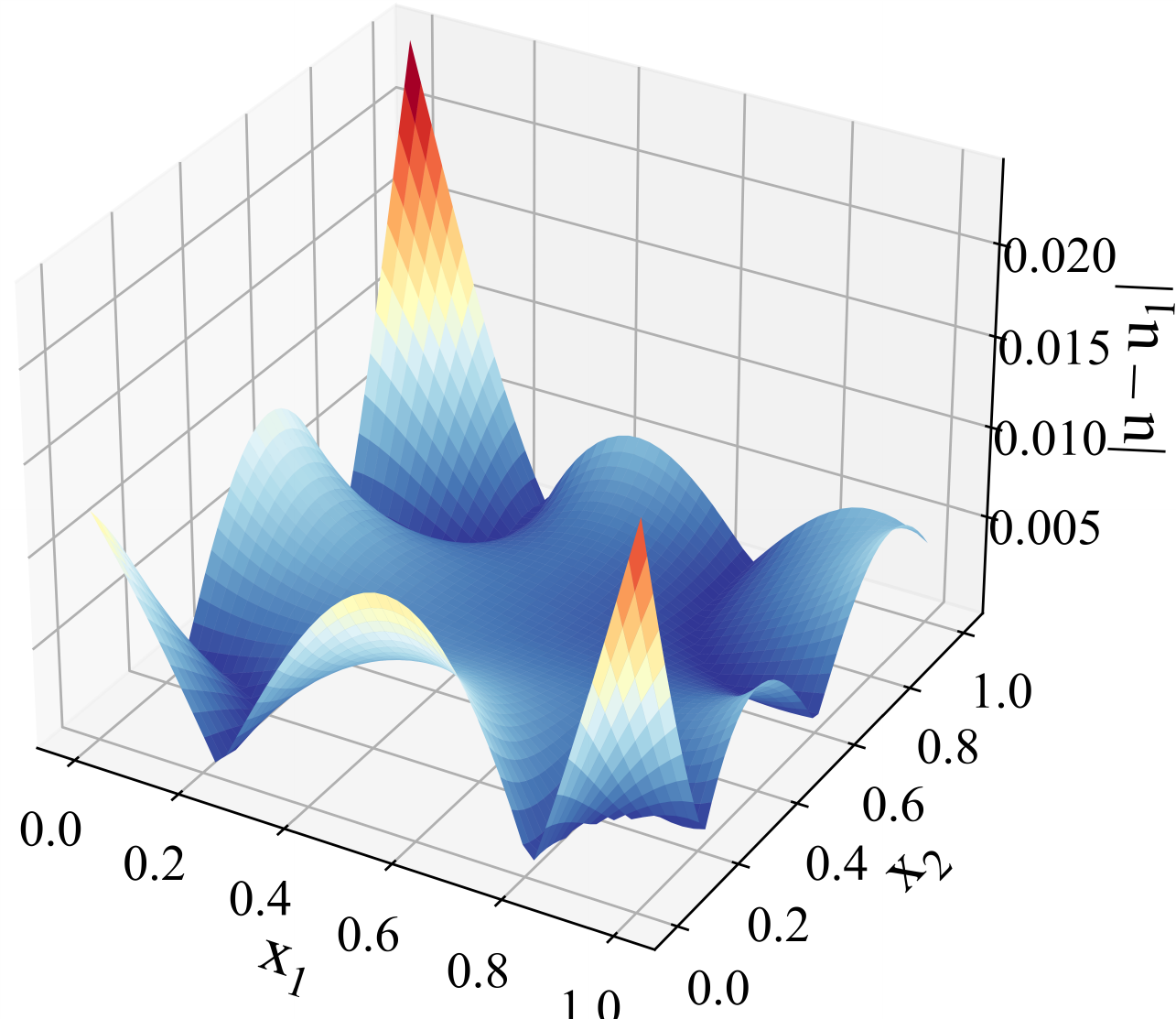}
    }    
\vspace{-2mm}
\caption{\textbf{Performance on 3D 4th-order Heat equation.} (a) The true solution $u$ in interval $x_1,x_2\in[0, 1]\times[0,1]$ at $t=0.5$. (b) \M's solution $u_1$. (c) The Taylor expansion result $u_2$ for the network on the reference point $(t,x_1,x_2)=(0.5, 0.5, 0.5)$. (d) Approximation error $|u-u_1|$.}
\label{3D 4order}
\end{center}
\vskip -0.2in
\end{figure}

\vspace{0.5mm}
\noindent\textbf{3D function.\quad} Further, we use \M~ to solve the 4th-order PDE of a heat equation defined as
\begin{equation}
        u_t-\nabla^4u=\pi^2 sin(\pi x_1)sin(\pi x_2)(cos(\pi t)-4\pi^2 sin(\pi t)),
\label{heat equation}
\end{equation}
where $\nabla^4$ is the square of the Laplacian operator w.r.t. $x_1$ and $x_2$.  
Fig.~\ref{3D 4order}(a) is the true PDE solution, and Fig.~\ref{3D 4order}(b)(c) show \M's solution and its 10-order Taylor polynomial.
The small residue between the network outputs and ground truth in Fig.~\ref{3D 4order}(d) shows that \M~achieves high accuracy. 

In our experiments, we also observed that while \M~ can provide an approximate solution for PDEs, the quality of the expansion depends on the values of the network parameters. Specifically, the expansion can serve as a good replacement for the neural network when with small parameter values which is in line with the convergence analysis presented in Section \ref{convergence}.

\section{Conclusions}

Aiming at solving high-order PDEs effectively, we derive the high-order derivative rule of neural network for quick and accurate derivative calculation, adopt it to develop a neural-network-based PDE solver, and expand the final black-box neural network into an explicit Taylor polynomial. The convergence condition of the Taylor series is analyzed experimentally validated as well.

\M~ has built a simple and general framework to enable obtaining the approximate solution of PDEs quickly. Comprehensive experiments are conducted to verify the high approximation accuracy of Taylor series to the target neural network, and the high efficiency in calculating partial derivatives. We also validate the high performance of \M~ on multiple high-order PDEs, from 1D to 3D. Moreover, \M~ provides an interpretable understanding of the learned black-box neural network, and can also be potentially used to specify the function parameters if given the form of the latent PDE solution. 


In the future, in addition to raising the accuracy further, we would like to apply \M~ to some different directions/topics, e.g., explaining the working mechanism of neural networks, developing high-order optimization algorithms to accelerate network training. Moreover, we can get the derivatives between any nodes of a neural network, which might inspire lightweight network design.  
Interpreting and simplifying a network describing the physical field or industrial controller can also be considered.

\bibliographystyle{unsrt}
\bibliography{main}

\begin{thebibliography}{10}

\bibitem{cybenko1989approximation}
George Cybenko.
\newblock Approximation by superpositions of a sigmoidal function.
\newblock {\em Mathematics of control, signals and systems}, 2(4):303--314,
  1989.

\bibitem{hornik1989multilayer}
Kurt Hornik, Maxwell Stinchcombe, and Halbert White.
\newblock Multilayer feedforward networks are universal approximators.
\newblock {\em Neural networks}, 2(5):359--366, 1989.

\bibitem{chen1995universal}
Tianping Chen and Hong Chen.
\newblock Universal approximation to nonlinear operators by neural networks
  with arbitrary activation functions and its application to dynamical systems.
\newblock {\em IEEE Transactions on Neural Networks}, 6(4):911--917, 1995.

\bibitem{leshno1993multilayer}
Moshe Leshno, Vladimir~Ya Lin, Allan Pinkus, and Shimon Schocken.
\newblock Multilayer feedforward networks with a nonpolynomial activation
  function can approximate any function.
\newblock {\em Neural networks}, 6(6):861--867, 1993.

\bibitem{dissanayake1994neural}
MWMG Dissanayake and Nhan Phan-Thien.
\newblock Neural-network-based approximations for solving partial differential
  equations.
\newblock {\em communications in Numerical Methods in Engineering},
  10(3):195--201, 1994.

\bibitem{pde-find}
Samuel~H. Rudy, Steven~L. Brunton, Joshua~L. Proctor, and J.~Nathan Kutz.
\newblock Data-driven discovery of partial differential equations.
\newblock {\em Science Advances}, 3(4):e1602614, 2017.

\bibitem{yu2018deep}
Bing Yu et~al.
\newblock The deep ritz method: a deep learning-based numerical algorithm for
  solving variational problems.
\newblock {\em Communications in Mathematics and Statistics}, 6(1):1--12, 2018.

\bibitem{sirignano2018dgm}
Justin Sirignano and Konstantinos Spiliopoulos.
\newblock Dgm: A deep learning algorithm for solving partial differential
  equations.
\newblock {\em Journal of computational physics}, 375:1339--1364, 2018.

\bibitem{raissi2019physics}
Maziar Raissi, Paris Perdikaris, and George~E Karniadakis.
\newblock Physics-informed neural networks: A deep learning framework for
  solving forward and inverse problems involving nonlinear partial differential
  equations.
\newblock {\em Journal of Computational physics}, 378:686--707, 2019.

\bibitem{lu2021deepxde}
Lu~Lu, Xuhui Meng, Zhiping Mao, and George~Em Karniadakis.
\newblock Deepxde: A deep learning library for solving differential equations.
\newblock {\em SIAM Review}, 63(1):208--228, 2021.

\bibitem{wang2021understanding}
Sifan Wang, Yujun Teng, and Paris Perdikaris.
\newblock Understanding and mitigating gradient flow pathologies in
  physics-informed neural networks.
\newblock {\em SIAM Journal on Scientific Computing}, 43(5):A3055--A3081, 2021.

\bibitem{liu2021dual}
Dehao Liu and Yan Wang.
\newblock A dual-dimer method for training physics-constrained neural networks
  with minimax architecture.
\newblock {\em Neural Networks}, 136:112--125, 2021.

\bibitem{xiang2021self}
Zixue Xiang, Wei Peng, Xiaohu Zheng, Xiaoyu Zhao, and Wen Yao.
\newblock Self-adaptive loss balanced physics-informed neural networks for the
  incompressible navier-stokes equations.
\newblock {\em arXiv preprint arXiv:2104.06217}, 2021.

\bibitem{meng2020ppinn}
Xuhui Meng, Zhen Li, Dongkun Zhang, and George~Em Karniadakis.
\newblock Ppinn: Parareal physics-informed neural network for time-dependent
  pdes.
\newblock {\em Computer Methods in Applied Mechanics and Engineering},
  370:113250, 2020.

\bibitem{moseley2021finite}
Ben Moseley, Andrew Markham, and Tarje Nissen-Meyer.
\newblock Finite basis physics-informed neural networks (fbpinns): a scalable
  domain decomposition approach for solving differential equations.
\newblock {\em arXiv preprint arXiv:2107.07871}, 2021.

\bibitem{lyu2022mim}
Liyao Lyu, Zhen Zhang, Minxin Chen, and Jingrun Chen.
\newblock Mim: A deep mixed residual method for solving high-order partial
  differential equations.
\newblock {\em Journal of Computational Physics}, 452:110930, 2022.

\bibitem{lu2021learning}
Lu~Lu, Pengzhan Jin, Guofei Pang, Zhongqiang Zhang, and George~Em Karniadakis.
\newblock Learning nonlinear operators via deeponet based on the universal
  approximation theorem of operators.
\newblock {\em Nature Machine Intelligence}, 3(3):218--229, 2021.

\bibitem{paszke2017automatic}
Adam Paszke, Sam Gross, Soumith Chintala, Gregory Chanan, Edward Yang, Zachary
  DeVito, Zeming Lin, Alban Desmaison, Luca Antiga, and Adam Lerer.
\newblock Automatic differentiation in pytorch.
\newblock 2017.

\bibitem{kingma2014adam}
Diederik~P Kingma and Jimmy Ba.
\newblock Adam: A method for stochastic optimization.
\newblock {\em arXiv preprint arXiv:1412.6980}, 2014.

\bibitem{sitzmann2020implicit}
Vincent Sitzmann, Julien Martel, Alexander Bergman, David Lindell, and Gordon
  Wetzstein.
\newblock Implicit neural representations with periodic activation functions.
\newblock {\em Advances in Neural Information Processing Systems},
  33:7462--7473, 2020.

\end{thebibliography}

\newpage
\appendix

\section{High-order derivatives of composite function}

Considering two functions $g(x)$ and $f(y)$, being $n$-order derivable at $x_0$ and $y_0=g(x_0)$ respectively. 
$\frac{\partial^k g}{\partial x^k}|_{x=x_0}$ and $\frac{\partial^k f}{\partial y^k}|_{y=y_0}$ are the $k$-order derivative of $g(x)$ at $x_0$ and of $f(y)$ at $y_0$. According to the chain rule, we can calculate the first three terms of $f(g(x))$'s $n$-order derivatives as
\begin{equation}  
\begin{split}
\left\{  
\begin{array}{lc}
    \frac{\partial f}{\partial x}
    =\frac{\partial g}{\partial x}\frac{\partial f}{\partial g},  \\    
    \frac{{\partial}^2 f}{\partial x^2}
    =\frac{\partial^2 g}{\partial x^2}\frac{\partial f}{\partial g}+(\frac{\partial g}{\partial x})^2\frac{\partial^2 f}{\partial g^2}, \\
    \frac{{\partial}^3 f}{\partial x^3}
    =\frac{\partial^3 g}{\partial x^3}\frac{\partial f}{\partial g}+3\frac{\partial g}{\partial x}\frac{\partial^2 g}{\partial x^2}\frac{\partial^2 f}{\partial g^2}+(\frac{\partial g}{\partial x})^3\frac{\partial^3 f}{\partial g^3}.
\end{array}
\right.  
\label{3-order derivatives sup}
\end{split}
\end{equation} 
For more terms, we convert $\frac{\partial }{\partial x}\frac{{\partial}^i f}{\partial g^i}$ to $\frac{\partial g}{\partial x}\frac{{\partial}^{i+1} f}{\partial g^{i+1}}$, and 
$\frac{{\partial}^n f}{\partial x^n}$ can be calculated given $\{\frac{\partial^i f}{\partial g^i}, i=1,\ldots,n\}$ and $\{\frac{\partial^i g}{\partial x^i}, i=1,\ldots,n\}$. Then Eq.~(\ref{3-order derivatives sup}) turns into following matrix form
\begin{equation}
\begin{split}
\left[
    \begin{array}{c}
            \frac{\partial f}{\partial x}  \\
            \vdots \\
            \frac{\partial^n f}{\partial x^n}
    \end{array}
\right]=
\left[
    \begin{array}{cccc}
            \frac{\partial g}{\partial x} & 0 & 0 & 0  \\
             \frac{\partial^2 g}{\partial x^2} & (\frac{\partial g}{\partial x})^2 & 0 & 0 \\
             \frac{\partial^3 g}{\partial x^3} & 3\frac{\partial g}{\partial x}\frac{\partial^2 g}{\partial x^2} & (\frac{\partial g}{\partial x})^3 & 0 \\
            \vdots & \vdots & \vdots & \ddots
    \end{array}
\right]
\left[
    \begin{array}{c}
            \frac{\partial f}{\partial g}  \\
            \vdots \\
            \frac{\partial^n f}{\partial g^n}
    \end{array}
\right],
\end{split}
\label{chain matrix sup}
\end{equation}
which can be further abbreviated as
\begin{equation}
    \PD^{f,x}=\CT^{g,x}\PD^{f,g}.
\label{abbreviated chain matrix sup}
\end{equation}
In this equation $\PD^{f,x} \in \mathbb{R}^n$ and $\PD^{f,g} \in \mathbb{R}^n$ are respectively the vectors composed of partial derivatives $\{\frac{\partial^i f}{\partial x^i}\}$ and  $\{\frac{\partial^i f}{\partial g^i}\}$; $\CT^{g,x} \in \mathbb{R}^{n\times n}$ is the chain transformation matrix composed of 
$\frac{\partial^i g}{\partial x^i}$ and takes a lower triangular form. So far, the calculation of $f(g(x))$'s $n$-order derivatives turns into the computation of $\CT^{g,x} \in \mathbb{R}^{n\times n}$.

From Eq. (\ref{chain matrix sup}) the $i$th term is 
\begin{equation}
    \frac{\partial^i f}{\partial x^i}=\sum_{j=1}^{n}\CT^{g,x}_{i,j}\frac{\partial^j f}{\partial g^j} (i<n),
\end{equation}
and we can derive $(i+1)$th term as
\begin{equation}
\begin{split}
    \frac{\partial^{i+1} f}{\partial x^{i+1}}
    &=\frac{\partial }{\partial x}\frac{\partial^i f}{\partial x^i}
    =\sum_{j=1}^{n}\frac{\partial }{\partial x}(\CT^{g,x}_{i,j}\frac{\partial^j f}{\partial g^j}) 
    =\sum_{j=1}^{n}\frac{\partial \CT^{g,x}_{i,j}}{\partial x}\frac{\partial^j f}{\partial g^j}+\sum_{j=1}^{n}\frac{\partial g}{\partial x}\CT^{g,x}_{i,j}\frac{\partial^{j+1} f}{\partial g^{j+1}} \\
    &=\sum_{j=1}^{n}\left(\frac{\partial \CT^{g,x}_{i,j}}{\partial x}
    +\frac{\partial g}{\partial x}\CT^{g,x}_{i,j-1} \right)\frac{\partial^j f}{\partial g^j}
    -\frac{\partial g}{\partial x}\CT^{g,x}_{i,0}\frac{\partial f}{\partial g}
    +\frac{\partial g}{\partial x}\CT^{g,x}_{i,n}\frac{\partial^{n+1} f}{\partial g^{n+1}}.
\end{split}
\label{eq:i+1th}
\end{equation}
Because $\CT^{g,x}_{i,0}=0$ and $\CT^{g,x}_{i,n}=0 (i < n)$, Eq.~(\ref{eq:i+1th}) can be simplified into
\begin{equation}
    \frac{\partial^{i+1} f}{\partial x^{i+1}}
    =\sum_{j=1}^{n}\left(\frac{\partial \CT^{g,x}_{i,j}}{\partial x}
    +\frac{\partial g}{\partial x}\CT^{g,x}_{i,j-1} \right)\frac{\partial^j f}{\partial g^j}.
\end{equation}
Therefore, the recurrence formula of  $\CT^{g,x}$ is
\begin{equation}  
\left\{  
    \begin{array}{lc}
        \CT^{g,x}_{1,1}=\frac{\partial g}{\partial x}  \\  
        \CT^{g,x}_{i,j}=0, i < j\\  
        \CT^{g,x}_{i+1,j}=\frac{\partial \CT^{g,x}_{i,j}}{\partial x}
         +\frac{\partial g}{\partial x}\CT^{g,x}_{i,j-1},  
    \end{array}  
\right.
\label{recurrence sup}
\end{equation} 
which explicitly composes 
the $n$-order chain transformation matrix $\CT^{g,x}$ in Eq.~(\ref{abbreviated chain matrix sup}).

\section{Matrix form of Taylor expansion}

To facilitate derivation,  for column vectors $\mathbf{e}=[e_1 \ldots e_u]^T$ and $\mathbf{r}=[r_1 \ldots r_v]^T$ we define an operator $\beta$ to save the information of their $k$-order partial derivatives, and following equations hold, 
\begin{equation}  
\f{\beta^k \B r^T}{\beta \B e^k}
=\left[
    \begin{array}{ccc}
        \f{\p^k\B r_1}{\p\B e_1^k} & \ldots & \f{\p^k\B r_v}{\p\B e_1^k} \\
        \vdots & \ddots & \vdots \\
        \f{\p^k\B r_1}{\p\B e_u^k} & \ldots & \f{\p^k\B r_v}{\p\B e_u^k}
    \end{array}
\right].
\end{equation} 

Based on above definition, Eq. (\ref{chain matrix sup}) can be re-written in matrix form as 
\begin{equation}
    \begin{split}
        \PD^{\mathbf{y},\mathbf{y}^{(m)}}=
    \left[
    \begin{array}{c}
            \frac{\beta \mathbf{y}}{\beta \mathbf{y}^{(m)}} \\
            \vdots \\
            \frac{\beta^n \mathbf{y}}{\beta {\mathbf{y}^{(m)}}^n}
    \end{array}
    \right],
    \end{split}
\label{PD matrix}
\end{equation}
and
\begin{equation}
    \begin{split}
        \CT^{\mathbf{y}^{(m+1)},\mathbf{y}^{(m)}}=
        \left[
        \begin{array}{ccc}
            \frac{\beta {\mathbf{y}^{(m+1)}}^T}{\beta \mathbf{y}^{(m)}} & 0 & 0 \\
            \frac{\beta^2 {\mathbf{y}^{(m+1)}}^T}{\beta {\mathbf{y}^{(m)}}^2} & 
            {(\frac{\beta {\mathbf{y}^{(m+1)}}^T}{\beta \mathbf{y}^{(m)}})}^{\circ 2} & 0\\
            \vdots & \vdots & \ddots
        \end{array}
        \right],
    \end{split}
\label{CT matrix}
\end{equation}
where $\circ k$ is Hadamard power, $(\mathbf{A}^{\circ k})_{i,j}=\mathbf{A}_{i,j}^k$. The form of Eqns.~(\ref{PD matrix})(\ref{CT matrix}) are consistent with Eq.~(\ref{chain matrix sup}), only with scalar elements replaced by matrices, and the operations between matrices are Hadamard power $\circ k$ and Hadamard product $\odot$.
See Eq.~(\ref{vMv app}) for the motivation of above definition.

\subsection{Ouput Layer}
The final output $\mathbf{y}=\mathbf{y}^{(d)}=\left[y_1^{d}\right]\in \mathbb{R}$, the derivatives are:
\begin{equation}
    \frac{\partial^k \mathbf{y}}{\partial {y_1^{(d)}}^k}=
    \left\{ 
    \begin{array}{lc}
        1, & k=1  \qquad \quad\\  
        0, & k=2,\ldots,n 
    \end{array}  
    \right.
\end{equation}

\begin{equation}
\frac{\beta^k \mathbf{y}}{\beta {\mathbf{y}^{(d)}}^k}=\left[\frac{\partial^k \mathbf{y}}{\partial {y_1^{(d)}}^k}\right]=
    \left\{ 
    \begin{array}{lc}
        \left[1\right], & k=1  \qquad \quad \\  
        \left[0\right], & k=2,\ldots,n 
    \end{array}  
    \right.
\end{equation}

\begin{equation}
    \begin{split}
    \PD^{\mathbf{y},\mathbf{y}^{(d)}}
    =\left[
    \begin{array}{c}
            \frac{\beta \mathbf{y}}{\beta \mathbf{y}^{(d)}} \\
            \vdots \\           
            \frac{\beta^n \mathbf{y}}{\beta {\mathbf{y}^{(d)}}^n}
    \end{array}
    \right]
    =\left[
    \begin{array}{c}
            1 \\
            0 \\
            \vdots \\           
            0
    \end{array}
    \right] \in \mathbb{R}^{n}.
    \end{split}
\label{Vd ap}
\end{equation}

\subsection{Hidden Layer}
\begin{equation}
\begin{split}
    \frac{\beta \mathbf{y}}{\beta {\mathbf{y}}^{(m)}}
    &=\left[
    \begin{array}{c}
        \frac{\partial \mathbf{y}}{\partial \mathbf{y}_1^{(m)}} \\
        \vdots \\
        \frac{\partial \mathbf{y}}{\partial \mathbf{y}_{o_m}^{(m)}}
    \end{array}
    \right] \\
    &=\left[
    \begin{array}{ccc}
        \frac{\partial {y}_1^{(m+1)}}{\partial {y}_1^{(m)}} &
        \ldots &
        \frac{\partial {y}_{o_{m+1}}^{(m+1)}}{\partial {y}_1^{(m)}} \\
        \vdots & \ddots & \vdots \\
        \frac{\partial {y}_1^{(m+1)}}{\partial {y}_{o_m}^{(m)}} &
        \ldots &
        \frac{\partial {y}_{o_{m+1}}^{(m+1)}}{\partial {y}_{o_m}^{(m)}}
    \end{array}
    \right]
    \left[
    \begin{array}{c}
        \frac{\partial \mathbf{y}}{\partial {y}_1^{(m+1)}} \\
        \vdots \\
        \frac{\partial \mathbf{y}}{\partial {y}_{o_{m+1}}^{(m+1)}}
    \end{array}
    \right] \\
    &=\frac{\beta {\mathbf{y}^{(m+1)}}^T}{\beta \mathbf{y}^{(m)}}
    \frac{\beta \mathbf{y}}{\beta \mathbf{y}^{(m+1)}}, \\
    \frac{\beta^2 \mathbf{y}}{\beta {\mathbf{y}^{(m)}}^2}
    &=\left[
    \begin{array}{c}
        \frac{\partial^2 \mathbf{y}}{\partial {\mathbf{y}_1^{(m)}}^2} \\
        \vdots \\
        \frac{\partial^2 \mathbf{y}}{\partial {\mathbf{y}_{o_m}^{(m)}}^2}
    \end{array}
    \right] \\
    &=\left[
    \begin{array}{ccc}
        \frac{\partial^2 {y}_1^{(m+1)}}{\partial {y_1^{(m)}}^2} &
        \ldots &
        \frac{\partial^2 y_{o_{m+1}}^{(m+1)}}{\partial {y_1^{(m)}}^2} \\
        \vdots & \ddots & \vdots \\
        \frac{\partial^2 {y}_1^{(m+1)}}{\partial {y_{o_m}^{(m)}}^2} &
        \ldots &
        \frac{\partial^2 {y}_{o_{m+1}}^{(m+1)}}{\partial {y_{o_m}^{(m)}}^2}
    \end{array}
    \right]
    \left[
    \begin{array}{c}
        \frac{\partial \mathbf{y}}{\partial {y}_1^{(m+1)}} \\
        \vdots \\
        \frac{\partial \mathbf{y}}{\partial {y}_{o_{m+1}}^{(m+1)}}
    \end{array}
    \right] 
    +\left[
    \begin{array}{ccc}
        \left(\frac{\partial {y}_1^{(m+1)}}{\partial {y}_1^{(m)}}\right)^2 &
        \ldots &
        \left(\frac{\partial {y}_{o_{m+1}}^{(m+1)}}{\partial {y}_1^{(m)}}\right)^2 \\
        \vdots & \ddots & \vdots \\
        \left(\frac{\partial {y}_1^{(m+1)}}{\partial {y}_{o_m}^{(m)}}\right)^2 &
        \ldots &
        \left(\frac{\partial {y}_{o_{m+1}}^{(m+1)}}{\partial {y}_{o_m}^{(m)}}\right)^2
    \end{array}
    \right]
    \left[
    \begin{array}{c}
        \frac{\partial^2 \mathbf{y}}{\partial {y_1^{(m+1)}}^2} \\
        \vdots \\
        \frac{\partial^2 \mathbf{y}}{\partial {y_{o_{m+1}}^{(m+1)}}^2} 
    \end{array}
    \right] \\
    &=\frac{\beta^2 {\mathbf{y}^{(m+1)}}^T}{\beta {\mathbf{y}^{(m)}}^2}
    \frac{\beta \mathbf{y}}{\beta \mathbf{y}^{(m+1)}}
    +{\frac{\beta {\mathbf{y}^{(m+1)}}^T}{\beta \mathbf{y}^{(m)}}}^{\circ 2}
    \frac{\beta^2 \mathbf{y}}{\beta {\mathbf{y}^{(m+1)}}^2}, \\
    \frac{\beta^3 \mathbf{y}}{\beta {\mathbf{y}^{(m)}}^3}
    &=\left[
    \begin{array}{c}
        \frac{\partial^3 \mathbf{y}}{\partial {\mathbf{y}_1^{(m)}}^3} \\
        \vdots \\
        \frac{\partial^3 \mathbf{y}}{\partial {\mathbf{y}_{o_m}^{(m)}}^3}
    \end{array}
    \right] \\
    &=\frac{\beta^3 {\mathbf{y}^{(m+1)}}^T}{\beta {\mathbf{y}^{(m)}}^3}
    \frac{\beta \mathbf{y}}{\beta \mathbf{y}^{(m+1)}}
    +3\frac{\beta {\mathbf{y}^{(m+1)}}^T}{\beta \mathbf{y}^{(m)}} \odot
    \frac{\beta^2 {\mathbf{y}^{(m+1)}}^T}{\beta {\mathbf{y}^{(m)}}^2}
    \frac{\beta^2 \mathbf{y}}{\beta {\mathbf{y}^{(m+1)}}^2}
    +{\frac{\beta {\mathbf{y}^{(m+1)}}^T}{\beta \mathbf{y}^{(m)}}}^{\circ 3}
    \frac{\beta^3 \mathbf{y}}{\beta {\mathbf{y}^{(m+1)}}^3}.
\end{split}
\end{equation}

Convert it to a matrix form
\begin{equation}
\begin{split}
    \left[
    \begin{array}{c}
        \frac{\beta \mathbf{y}}{\beta {\mathbf{y}^{(m)}}} \\
        \frac{\beta^2 \mathbf{y}}{\beta {\mathbf{y}^{(m)}}^2} \\
        \frac{\beta^3 \mathbf{y}}{\beta {\mathbf{y}^{(m)}}^3} \\
        \vdots \\
        \frac{\beta^n \mathbf{y}}{\beta {\mathbf{y}^{(m)}}^n}
    \end{array}
    \right]
    =\left[
    \begin{array}{cccc}
        \frac{\beta {\mathbf{y}^{(m+1)}}^T}{\beta \mathbf{y}^{(m)}} & 0 & 0 & 0\\
        \frac{\beta^2 {\mathbf{y}^{(m+1)}}^T}{\beta {\mathbf{y}^{(m)}}^2} & 
        {(\frac{\beta {\mathbf{y}^{(m+1)}}^T}{\beta \mathbf{y}^{(m)}})}^{\circ 2} & 0 & 0\\
        \frac{\beta^3 {\mathbf y^{(m+1)}}^T}{\beta {\mathbf y^{(m)}}^3} & 
        3\frac{\beta {\mathbf y^{(m+1)}}^T}{\beta \mathbf y^{(m)}}\odot \frac{\beta^2 {\mathbf y^{(m+1)}}^T}{\beta {\mathbf y^{(m)}}^2} &
        {(\frac{\beta {\mathbf y^{(m+1)}}^T}{\beta \mathbf y^{(m)}})}^{\circ 3} & 0\\
        \vdots & \vdots & \vdots & \ddots
    \end{array}
    \right]
    \left[
    \begin{array}{c}
        \frac{\beta \mathbf{y}}{\beta {\mathbf{y}^{(m+1)}}} \\
        \frac{\beta^2 \mathbf{y}}{\beta {\mathbf{y}^{(m+1)}}^2} \\
        \frac{\beta^3 \mathbf{y}}{\beta {\mathbf{y}^{(m+1)}}^3} \\
        \vdots \\
        \frac{\beta^n \mathbf{y}}{\beta {\mathbf{y}^{(m+1)}}^n}
    \end{array}
    \right]
\end{split}
\label{vMv app}
\end{equation}

The form of the above formula is consistent with Eq.~(\ref{chain matrix sup}), only scalar elements are replaced by matrices, and the operations between matrices are element-wise power $\circ k$ and element-wise product $\odot$.
To simplified expression, the above equation is further abbreviated as

\begin{equation}
\begin{split}
    \PD^{\mathbf{y},\mathbf{y}^{m}}=
    \CT^{\mathbf{y}^{(m+1)},\mathbf{y}^{(m)}}
    \PD^{\mathbf{y},\mathbf{y}^{m+1}},
\end{split}
\end{equation}

and that's the reason for our definition for Eqns.~(\ref{PD matrix})(\ref{CT matrix}).

To simplify the expression, we omit the superscripts of $\PD^{\mathbf y,\mathbf y^{(m)}}$ and $\CT^{\mathbf y^{(m+1)},\mathbf y^{(m)}}$ as $\PD_{m}$ and $\CT_{m+1}$ respectively. Setting $\PD_{0}=\PD^{\mathbf y,\mathbf x}$ and $\CT_{1}=\CT^{\mathbf y^{(1)},\mathbf x}$. We have know $\PD^{\mathbf{y},\mathbf{y}^{d}}$ from Eq.~(\ref{Vd ap}) and the above formula, then we can get
\begin{eqnarray}
        \PD_{d}&=
        &\left[
        \begin{array}{cccc}
            1 & 0 & \ldots & 0
        \end{array}
        \right]^T, \label{vd app}
    \\
        \PD_{m}&=&\CT_{m+1}\PD_{m+1}, m=0,\ldots,d-1, \label{vm app}
\end{eqnarray}

in which $\PD_{m}$ contains all the $n$-order partial differential of a single variable $\{\frac{\partial^k \mathbf{y}}{\partial {y_i^{(m)}}^k}: k=1,\ldots,n; i=1,\ldots,o_m\}$.

When we know all the $\frac{\beta^k {\mathbf{y}^{(m+1)}}^T}{\beta {\mathbf{y}^{(m)}}^k}$, we can calculate $\CT_{m+1}$ with Eq.~(\ref{chain matrix sup}). So far, the calculation of $\PD_{m}$ turns into the computation of $\frac{\beta^k {\mathbf{y}^{(m+1)}}^T}{\beta {\mathbf{y}^{(m)}}^k}$.


\begin{equation}
\begin{split}
    \frac{\beta^k {\mathbf{y}^{(m+1)}}^T}{\beta {\mathbf{y}^{(m)}}^k}
    &=\left[
    \begin{array}{ccc}
        \frac{\partial^k y_1^{(m+1)}}{\partial {y_1^{(m)}}^k} & \ldots & \frac{\partial^k y_{o_{m+1}}^{(m+1)}}{\partial {y_1^{(m)}}^k} \\
         \vdots & \ddots & \vdots \\
         \frac{\partial^k y_1^{(m+1)}}{\partial {y_{o_m}^{(m)}}^k} & \ldots & \frac{\partial^k y_{o_{m+1}}^{(m+1)}}{\partial {y_{o_m}^{(m)}}^k} 
    \end{array}
    \right] \\
    &=\left[
    \begin{array}{ccc}
        {\mathbf{W}_{1,1}^{(m+1)}}^k
        \frac{\partial^k \sigma ({x}_{1}^{(m+1)})}{\partial {{x}_{1}^{(m+1)}}^k}
        & \ldots &
        {\mathbf{W}_{o_{m+1},1}^{(m+1)}}^k
        \frac{\partial^k \sigma ({x}_{o_{m+1}}^{(m+1)})}{\partial {{x}_{o_{m+1}}^{(m+1)}}^k}
        \\
        \vdots & \ddots & \vdots \\
        {\mathbf{W}_{1,o_m}^{(m+1)}}^k
        \frac{\partial^k \sigma ({x}_{1}^{(m+1)})}{\partial {{x}_{1}^{(m+1)}}^k}
        & \ldots &
        {\mathbf{W}_{o_{m+1},o_m}^{(m+1)}}^k
        \frac{\partial^k \sigma ({x}_{o_{m+1}}^{(m+1)})}{\partial {{x}_{o_{m+1}}^{(m+1)}}^k}
    \end{array}
    \right] \\
    &={\left({\mathbf{W}^{(m+1)}}^T\right)}^{\circ k} \odot 
    \left[
    \begin{array}{ccc}
        \frac{\partial^k \sigma ({x}_{1}^{(m+1)})}{\partial {{x}_{1}^{(m+1)}}^k}
        & \ldots &
        \frac{\partial^k \sigma ({x}_{o_{m+1}}^{(m+1)})}{\partial {{x}_{o_{m+1}}^{(m+1)}}^k}
        \\
        \vdots & \ddots & \vdots \\
        \frac{\partial^k \sigma ({x}_{1}^{(m+1)})}{\partial {{x}_{1}^{(m+1)}}^k}
        & \ldots &
        \frac{\partial^k \sigma ({x}_{o_{m+1}}^{(m+1)})}{\partial {{x}_{o_{m+1}}^{(m+1)}}^k}
    \end{array}
    \right] \\
    &=
    {\left({\mathbf{W}^{(m+1)}}^T\right)}^{\circ k} \odot 
    \left(
    \mathbf{1}_{o_m} \otimes 
    \left[
    \begin{array}{ccc}
        \frac{\partial^k \sigma ({x}_{1}^{(m+1)})}{\partial {{x}_{1}^{(m+1)}}^k}
        & \ldots &
        \frac{\partial^k \sigma ({x}_{o_{m+1}}^{(m+1)})}{\partial {{x}_{o_{m+1}}^{(m+1)}}^k}
    \end{array}
    \right]
    \right) \\
    &={\left({\mathbf{W}^{(m+1)}}^T\right)}^{\circ k} \odot 
    \left[
    \mathbf{1}_{o_m} \otimes \left(
    \frac{\partial^k \sigma (\mathbf x^{(m+1)})}{\partial {\mathbf x^{(m+1)}}^k}
    \right)^T
    \right]
\end{split}
\label{beta app}
\end{equation}

where $\mathbf{1}_{o_m} \in \mathbb{R}^{o_m}$ is an all-1 column vector, $\frac{\partial^k \sigma (\mathbf x^{(m+1)})}{\partial {\mathbf x^{(m+1)}}^k}=\left[\frac{\partial^k \sigma ({x}_{1}^{(m+1)})}{\partial {{x}_{1}^{(m+1)}}^k}, \ldots, \frac{\partial^k \sigma ({x}_{o_{m+1}}^{(m+1)})}{\partial {{x}_{o_{m+1}}^{(m+1)}}^k}\right]^T\in \mathbb{R}^{o_{m+1}}$ is $k$-order derivative of activation function. 

\subsection{Input layer}
If the mixed partial derivatives are not required, Eqns.~(\ref{vd app})(\ref{vm app}) are enough for obtaining $\PD_{0}$, which contains all the $n$-order partial differential of a single variable $\{\frac{\partial^k \mathbf{y}}{\partial x_i^k}: k=1,\ldots,n; i=1,\ldots,p\}$.

Similar to $\beta$, we define operator $\gamma$ to save all the mixed partial derivatives
\begin{equation}  
\left\{  
    \begin{array}{lc}
    \frac{\gamma \mathbf{y}}{\gamma \mathbf{x}}\in \mathbb{R}^{p}, (\frac{\gamma \mathbf{y}}{\gamma \mathbf{x}})_{i}=\frac{\partial \mathbf{y}}{\partial {x}_i}, &  \\  
    \frac{\gamma }{\gamma \mathbf{x}}\in \mathbb{R}^{p}, (\frac{\gamma }{\gamma \mathbf{x}})_{i}=\frac{\partial }{\partial {x}_i}, & \\  
     \frac{\gamma^k \mathbf{y}}{\gamma \mathbf{x}^k}=\frac{\gamma }{\gamma \mathbf{x}}\otimes \frac{\gamma^{k-1} \mathbf{y}}{\gamma \mathbf{x}^{k-1}} \in \mathbb{R}^{p^k},&
    \end{array}  
\right.  
\end{equation} 
where $\frac{\gamma }{\gamma \mathbf{x}}\in \mathbb{R}^{p}$ is a column vector containing partial derivative operators, and $\frac{\gamma^k \mathbf{y}}{\gamma \mathbf{x}^k}$ contains all $k$-order partial derivatives.

\begin{equation}
\begin{split}
    \frac{\partial \mathbf{y}}{\partial x_j}
    &=\sum_{i=1}^{o_1}\frac{\partial y_i^{(1)}}{\partial x_j}\frac{\partial \mathbf{y}}{\partial y_i^{(1)}}
    =\sum_{i=1}^{o_1}\mathbf{W}_{i,j}^{(1)}\frac{\partial \sigma (x_i^{(1)})}{\partial x_i^{(1)}}\frac{\partial \mathbf{y}}{\partial y_i^{(1)}}
    =\left[
    \begin{array}{ccc}
       \mathbf{W}_{1,j}^{(1)} \frac{\partial \sigma (x_1^{(1)})}{\partial x_1^{(1)}} 
       & \ldots 
       & \mathbf{W}_{o_1,j}^{(1)} \frac{\partial \sigma (x_{o_1}^{(1)})}{\partial x_{o_1}^{(1)}}
    \end{array}
    \right]
    \left[
    \begin{array}{c}
       \frac{\partial \mathbf{y}}{\partial y_1^{(1)}} \\
       \vdots \\
       \frac{\partial \mathbf{y}}{\partial y_{o_1}^{(1)}}
    \end{array}
    \right] \\
    &={\mathbf{W}_{:,j}^{(1)}}^T\odot 
    \frac{\partial^k \sigma (\mathbf x^{(1)})}{\partial {\mathbf x^{(1)}}^k}^T
    \frac{\beta \mathbf{y}}{\beta \mathbf{y}^{(1)}},
\end{split}
\end{equation}

\begin{equation}
\begin{split}
    \frac{\gamma \mathbf{y}}{\gamma \mathbf{x}}
    &=\left[
    \begin{array}{c}
        \frac{\partial \mathbf{y}}{\partial x_1} \\
        \vdots \\
        \frac{\partial \mathbf{y}}{\partial x_p}
    \end{array}
    \right]
    =\left[
    \begin{array}{c}
        {\mathbf{W}_{:,1}^{(1)}}^T \\
        \vdots \\
        {\mathbf{W}_{:,p}^{(1)}}^T
    \end{array}
    \right]
    \odot
    \left[
    \begin{array}{c}
        \left(\frac{\partial \sigma (\mathbf x^{(1)})}{\partial {\mathbf x^{(1)}}}\right)^T \\
        \vdots \\
        \left(\frac{\partial \sigma (\mathbf x^{(1)})}{\partial {\mathbf x^{(1)}}}\right)^T
    \end{array}
    \right]
    \left[
        \frac{\beta \mathbf{y}}{\beta \mathbf{y}^{(1)}}
    \right]
    ={\mathbf{W}^{(1)}}^T \odot
        \left[
        \mathbf{1}_{p} \otimes
        \left(
        \frac{\partial \sigma (\mathbf x^{(1)})}{\partial {\mathbf x^{(1)}}}
        \right)^T
        \right]
        \left[
        \frac{\beta \mathbf{y}}{\beta \mathbf{y}^{(1)}}
    \right],
\end{split}
\end{equation}

\begin{equation}
\begin{split}
    \frac{\partial^2 \mathbf{y}}{\partial x_ix_j}=
    \left[
    \begin{array}{cc}
        {\mathbf{W}_{:,i}^{(1)}}^T \odot {\mathbf{W}_{:,j}^{(1)}}^T &
        {\mathbf{W}_{:,i}^{(1)}}^T \odot {\mathbf{W}_{:,j}^{(1)}}^T
    \end{array}
    \right]
    \odot
    \left[
    \begin{array}{cc}
        \frac{\partial^2 \sigma (\mathbf x^{(1)})}{\partial {\mathbf x^{(1)}}^2}^T &
        \left(\frac{\partial \sigma (\mathbf x^{(1)})}{\partial {\mathbf x^{(1)}}}^T\right)^{\circ 2}
    \end{array}
    \right]
    \left[
    \begin{array}{c}
        \frac{\beta \mathbf{y}}{\beta \mathbf{y}^{(1)}} \\
        \frac{\beta^2 \mathbf{y}^T}{\beta {\mathbf{y}^{(1)}}^2}
    \end{array}
    \right] \qquad \qquad,
\end{split}
\end{equation}

\begin{equation}
\begin{split}
    &\frac{\gamma^2 \mathbf{y}}{\gamma \mathbf{x}^2}
    =\frac{\gamma }{\gamma \mathbf{x}} \otimes 
    \frac{\gamma \mathbf{y}}{\gamma \mathbf{x}}
    =\left[
    \begin{array}{c}
        \frac{\partial^2 \mathbf{y}}{\partial x_1^2},
        \cdots,
        \frac{\partial^2 \mathbf{y}}{\partial x_1x_p},
        \frac{\partial^2 \mathbf{y}}{\partial x_2x_1},
        \cdots,
        \frac{\partial^2 \mathbf{y}}{\partial x_2x_p},
        \cdots,
        \frac{\partial^2 \mathbf{y}}{\partial x_px_1},
        \cdots,
        \frac{\partial^2 \mathbf{y}}{\partial x_p^2}
    \end{array}
    \right]^T \\
    &=\left[
    \begin{array}{cc}
        {\mathbf{W}_{:,1}^{(1)}}^T \odot {\mathbf{W}_{:,1}^{(1)}}^T &
        {\mathbf{W}_{:,1}^{(1)}}^T \odot {\mathbf{W}_{:,1}^{(1)}}^T \\
        \vdots & \vdots\\
        {\mathbf{W}_{:,1}^{(1)}}^T \odot {\mathbf{W}_{:,p}^{(1)}}^T &
        {\mathbf{W}_{:,1}^{(1)}}^T \odot {\mathbf{W}_{:,p}^{(1)}}^T \\
        {\mathbf{W}_{:,2}^{(1)}}^T \odot {\mathbf{W}_{:,1}^{(1)}}^T &
        {\mathbf{W}_{:,2}^{(1)}}^T \odot {\mathbf{W}_{:,1}^{(1)}}^T \\
        \vdots & \vdots\\
        {\mathbf{W}_{:,2}^{(1)}}^T \odot {\mathbf{W}_{:,p}^{(1)}}^T &
        {\mathbf{W}_{:,2}^{(1)}}^T \odot {\mathbf{W}_{:,p}^{(1)}}^T \\
        \vdots & \vdots\\
        {\mathbf{W}_{:,p}^{(1)}}^T \odot {\mathbf{W}_{:,1}^{(1)}}^T &
        {\mathbf{W}_{:,p}^{(1)}}^T \odot {\mathbf{W}_{:,1}^{(1)}}^T \\
        \vdots & \vdots\\
        {\mathbf{W}_{:,p}^{(1)}}^T \odot {\mathbf{W}_{:,p}^{(1)}}^T &
        {\mathbf{W}_{:,p}^{(1)}}^T \odot {\mathbf{W}_{:,p}^{(1)}}^T
    \end{array}
    \right]
    \odot
    \left[
    \begin{array}{cc}
        \frac{\partial^2 \sigma (\mathbf x^{(1)})}{\partial {\mathbf x^{(1)}}^2}^T &
        \left(\frac{\partial \sigma (\mathbf x^{(1)})}{\partial {\mathbf x^{(1)}}}^T\right)^{\circ 2} \\
        \vdots & \vdots \\
        \frac{\partial^2 \sigma (\mathbf x^{(1)})}{\partial {\mathbf x^{(1)}}^2}^T &
        \left(\frac{\partial \sigma (\mathbf x^{(1)})}{\partial {\mathbf x^{(1)}}}^T\right)^{\circ 2} \\
        \frac{\partial^2 \sigma (\mathbf x^{(1)})}{\partial {\mathbf x^{(1)}}^2}^T &
        \left(\frac{\partial \sigma (\mathbf x^{(1)})}{\partial {\mathbf x^{(1)}}}^T\right)^{\circ 2} \\
        \vdots & \vdots \\
        \frac{\partial^2 \sigma (\mathbf x^{(1)})}{\partial {\mathbf x^{(1)}}^2}^T &
        \left(\frac{\partial \sigma (\mathbf x^{(1)})}{\partial {\mathbf x^{(1)}}}^T\right)^{\circ 2} \\
        \vdots & \vdots \\
        \frac{\partial^2 \sigma (\mathbf x^{(1)})}{\partial {\mathbf x^{(1)}}^2}^T &
        \left(\frac{\partial \sigma (\mathbf x^{(1)})}{\partial {\mathbf x^{(1)}}}^T\right)^{\circ 2} \\
        \vdots & \vdots \\
        \frac{\partial^2 \sigma (\mathbf x^{(1)})}{\partial {\mathbf x^{(1)}}^2}^T &
        \left(\frac{\partial \sigma (\mathbf x^{(1)})}{\partial {\mathbf x^{(1)}}}^T\right)^{\circ 2} \\
    \end{array}
    \right]
    \left[
    \begin{array}{c}
        \frac{\beta \mathbf{y}}{\beta \mathbf{y}^{(1)}} \\
        \frac{\beta^2 \mathbf{y}^T}{\beta {\mathbf{y}^{(1)}}^2}
    \end{array}
    \right] \\
    &=\left[
        \begin{array}{cc}
        \left({\mathbf{W}^{(1)}}^T \otimes \mathbf{1}_p\right) \odot 
        \left(\mathbf{1}_p \otimes {\mathbf{W}^{(1)}}^T\right) & 
        \left({\mathbf{W}^{(1)}}^T \otimes \mathbf{1}_p\right) \odot 
        \left(\mathbf{1}_p \otimes {\mathbf{W}^{\{ 1\}}}^T\right)
        \end{array}
    \right] 
    \odot \\
    &\quad\left[
        \begin{array}{cc}
        \mathbf{1}_{p^2}\otimes \frac{\partial^2 \sigma (\mathbf x^{(1)})}{\partial {\mathbf x^{(1)}}^2}^T &
        \left(\mathbf{1}_{p^2}\otimes \frac{\partial \sigma (\mathbf x^{(1)})}{\partial {\mathbf x^{(1)}}}^T\right)^{\circ 2}
        \end{array}
    \right]
    \left[
    \begin{array}{c}
        \frac{\beta \mathbf{y}}{\beta \mathbf{y}^{(1)}} \\
        \frac{\beta^2 \mathbf{y}^T}{\beta {\mathbf{y}^{(1)}}^2}
    \end{array}
    \right].
\end{split}
\end{equation}

By analogy, we can get a simple formula
\begin{equation}
\begin{split}
    \left[
    \begin{array}{c}
            \frac{\gamma \mathbf{y}}{\gamma \mathbf{x}}  \\
            \vdots \\
            \frac{\gamma^n \mathbf{y}}{\gamma \mathbf{x}^n}
    \end{array}
    \right]
    \!=\!
    \left[
     \begin{array}{c}
        \mathbf{W}_{1} \\
        \mathbf{W}_{2} \\
        \vdots
    \end{array}
    \right]\!\odot\!
    \left[
     \begin{array}{ccc}
        \sigma_{1}(\mathbf{x}^{(1)}) & 0 & 0  \\
        \sigma_{2}(\mathbf{x}^{(1)}) & 
        \sigma_{1}(\mathbf{x}^{(1)})^{\circ 2} & 0 \\
        \vdots & \vdots & \ddots
    \end{array}
    \right]\PD_1,
\label{total derivatives}
\end{split}
\end{equation}
with
\begin{equation}
\begin{split}
    \mathbf{W}_0&=\mathbf{1}_{n\times o_1}^T, \\
    \mathbf{W}_{k}&=(\mathbf{1}_n^T \otimes {\mathbf{W}^{(1)}}^T \otimes \mathbf{1}_{p^{k-1}}) \odot(\mathbf{1}_{p}\otimes \mathbf{W}_{k-1}) \in \mathbb{R}^{p^k \times (n\times o_1)},
\label{Wk}
\end{split}
\end{equation}

and
\begin{equation}
    \sigma_{k}(\mathbf{x}^{(1)})=\mathbf{1}_{p^k}\otimes \left(\frac{\partial^k \sigma (\mathbf{x}^{(1)})}{\partial {\mathbf{x}^{(1)}}^k}\right)^T \in \mathbb{R}^{p^k \times o_1}.
\end{equation}

\section{High-order derivatives of nonlinear activation function}
\subsection{Sine}
$\sigma (x)=sin(x)$. The derivatives are
\begin{equation}
\begin{split}
    \frac{\partial^k \sigma (x)}{\partial x^k}=
    \left\{
    \begin{array}{lc}
    cos(x), & k \textit{ mod } 4=1 \\
    -sin(x), & k \textit{ mod } 4=2 \\
    -cos(x), & k \textit{ mod } 4=3 \\
    sin(x). & k \textit{ mod } 4=0 \\
    \end{array}
    \right.
\end{split}
\end{equation}

\subsection{ReLU}
$\sigma (x)=max(0,x)$. The derivatives are
\begin{equation}
\begin{split}
    \frac{\partial^k \sigma (x)}{\partial x^k}=
    \left\{
    \begin{array}{lc}
    1, & \text{if }k=1 \text{ and } x>0 \\
    0. & \text{else} \qquad \qquad \qquad
    \end{array}
    \right.
\end{split}
\end{equation}

\subsection{Sigmoid}
$\sigma (x)=\frac{e^x}{1+e^x}$. Abbreviate $\frac{\partial^k \sigma (x)}{\partial x^k}$ as $\sigma^{(k)} (x)$, and the first derivative is 

\begin{equation}
    \sigma^{(1)} (x)=\frac{e^x}{(1+e^x)^2}=\sigma (x)(1-\sigma (x))
    =\sigma (x)-\sigma (x)^2
\end{equation}
Note that $\sigma^{(1)} (x)=\sigma (x)-\sigma (x)^2$, we can express $\sigma^{(k)} (x)$ as the form containing only $\sigma (x)$.
\begin{equation}
\begin{split}
    \sigma^{(2)} (x)&=\sigma^{(1)} (x)-2\sigma (x)\sigma^{(1)} (x) \\
    &=[\sigma (x)-\sigma (x)^2]-2\sigma (x)[\sigma (x)-\sigma (x)^2] \\
    &=\sigma (x)-3\sigma (x)^2+2\sigma (x)^3.
\end{split}
\end{equation}

Organize it into matrix form:
\begin{equation}
\begin{split}
\left[
    \begin{array}{c}
            \sigma (x) \\
            \sigma^{(1)} (x) \\
            \sigma^{(2)} (x) \\
            \vdots \\
            \sigma^{(n)} (x)
    \end{array}
\right]=
\left[
    \begin{array}{cccc}
            1 & 0 & 0 & 0 \\
            1 & -1 & 0 & 0 \\
            1 & -3 & 2 & 0 \\
            \vdots & \vdots & \vdots & \ddots
    \end{array}
\right]
\left[
    \begin{array}{c}
            \sigma (x)  \\
            \sigma (x)^2  \\
            \sigma (x)^3  \\
            \vdots \\
            \sigma (x)^{n+1}
    \end{array}
\right].
\end{split}
\end{equation}
The square matrix is abbreviated as $B\in \mathbb{R}^{n+1\times n+1}$.

\begin{equation}
\begin{split}
    \sigma^{(k)} (x)=\sum_{i=1}^{k+1} B_{k+1,i}\sigma (x)^i,
\end{split}
\end{equation}

\begin{equation}
\begin{split}
    \sigma^{(k+1)} (x)&=\sum_{i=1}^{k+1} iB_{k+1,i}\sigma (x)^{i-1}\sigma^{(1)} (x) \\
    &=\sum_{i=1}^{k+1} iB_{k+1,i}\sigma (x)^{i-1}[\sigma (x)-\sigma (x)^2] \\
    &=\sum_{i=1}^{k+1} iB_{k+1,i}\sigma (x)^i-\sum_{i=1}^{k+1} iB_{k+1,i}\sigma (x)^{i+1} \\
    &=\sum_{i=1}^{k+1} iB_{k+1,i}\sigma (x)^i-\sum_{i=2}^{k+2} (i-1)B_{k+1,i-1}\sigma (x)^{i} \\
    &=B_{k+1,1}\sigma (x)+\sum_{i=2}^{k+1}[iB_{k+1,i}-(i-1)B_{k+1,i-1}]\sigma (x)^i-(k+1)B_{k+1,k+1}\sigma (x)^{k+2} \\
    &=\sum_{i=1}^{k+2} B_{k+2,i}\sigma (x)^i.
\end{split}
\end{equation}
Therefore, we can get a relationship:
\begin{equation}
\begin{split}
    B_{k+2,1}&=B_{k+1,1}, \\
    B_{k+2,i}&=iB_{k+1,i}-(i-1)B_{k+1,i-1}, i=2,\ldots,k+1 \\
    B_{k+2,k+2}&=-(k+1)B_{k+1,k+1}.
\end{split}
\end{equation}

The recurrence formula of  $B$ is
\begin{equation}  
\left\{  
    \begin{array}{lc}
        B_{i,j}=0, & i<j \qquad \qquad \quad \\
        B_{i,1}=1, & i=1,\ldots,n+1 \\
        B_{k,i}=iB_{k-1,i}-(i-1)B_{k-1,i-1}, & i=2,\ldots,k-1 \\
        B_{k,k}=-(k-1)B_{k-1,k-1}. & 
    \end{array}  
\right.
\label{recurrence2}
\end{equation} 

\subsection{Tanh}
$\sigma (x)=\frac{e^x-e^{-x}}{e^x+e^{-x}}$. The first two derivatives are
\begin{equation}
    \sigma^{(1)} (x)=1-(\frac{e^x-e^{-x}}{e^x+e^{-x}})^2=1-\sigma (x)^2.
\end{equation}

\begin{equation}
    \sigma^{(2)} (x)=-2\sigma (x)\sigma^{(1)} (x)=-2\sigma (x)+2\sigma (x)^3.
\end{equation}

Organize it into matrix form:
\begin{equation}
\begin{split}
\left[
    \begin{array}{c}
            1 \\
            \sigma (x) \\
            \sigma^{(1)} (x) \\
            \sigma^{(2)} (x) \\
            \vdots \\
            \sigma^{(n)} (x)
    \end{array}
\right]=
\left[
    \begin{array}{ccccc}
            1 & 0 & 0 & 0 & 0 \\
            0 & 1 & 0 & 0 & 0 \\
            1 & 0 & -1 & 0 & 0 \\
            0 & -2 & 0 & 2 & 0 \\
            \vdots & \vdots & \vdots & \vdots & \ddots
    \end{array}
\right]
\left[
    \begin{array}{c}
            1 \\
            \sigma (x)  \\
            \sigma (x)^2  \\
            \sigma (x)^3  \\
            \vdots \\
            \sigma (x)^{n+1}
    \end{array}
\right].
\end{split}
\end{equation}
The square matrix is abbreviated as $\mathbf{C}\in \mathbb{R}^{n+2\times n+2}$.
\begin{equation}
\begin{split}
    \sigma^{(k)} (x)=\sum_{i=1}^{k+2} C_{k+2,i}\sigma (x)^{i-1}, (k>=1)
\end{split}
\end{equation}

\begin{equation}
\begin{split}
    \sigma^{(k+1)} (x)&=\sum_{i=1}^{k+2} (i-1)C_{k+2,i}\sigma (x)^{i-2}\sigma^{(1)} (x) \\
    &=\sum_{i=1}^{k+2} (i-1)C_{k+2,i}\sigma (x)^{i-2}[1-\sigma (x)^2] \\
    &=\sum_{i=1}^{k+2} (i-1)C_{k+2,i}\sigma (x)^{i-2}-\sum_{i=1}^{k+2} (i-1)C_{k+2,i}\sigma (x)^{i} \\
    &=\sum_{i=0}^{k+1} iC_{k+2,i+1}\sigma (x)^{i-1}-\sum_{i=2}^{k+3} (i-2)C_{k+2,i-1}\sigma (x)^{i-1} \\
    &=C_{k+2,2}+\sum_{i=2}^{k+2}[iC_{k+2,i+1}-(i-2)C_{k+2,i-1}]\sigma (x)^{i-1}-(k+1)C_{k+2,k+2}\sigma (x)^{k+2} \\
    &=\sum_{i=1}^{k+3} C_{k+3,i}\sigma (x)^{i-1}.
\end{split}
\end{equation}
Therefore, we can get a relationship:

\begin{equation}
\begin{split}
    C_{k+3,1}&=C_{k+2,2}, \\
    C_{k+3,i}&=iC_{k+2,i+1}-(i-2)C_{k+2,i-1}, i=2,\ldots,k+2, \\
    C_{k+3,k+3}&=-(k+1)C_{k+2,k+2}.
\end{split}
\end{equation}

The recurrence formula of C is
\begin{equation}  
\left\{  
    \begin{array}{lc}
        C_{1,1}=1, C_{2,1}=0, C_{2,2}=1, & \\
        C_{i,j}=0, & i<j \qquad \qquad \quad \\
        C_{k,1}=C_{k-1,2}, & k=2,\ldots,n+2 \\
        C_{k,i}=iC_{k-1,i+1}-(i-2)C_{k-1,i-1}, & i=2,\ldots,k-1, \\
        C_{k,k}=-(k-2)C_{k-1,k-1}. & 
    \end{array}  
\right.
\label{recurrence3}
\end{equation} 

\subsection{No activation}
$\sigma (x)=x$. The derivatives are
\begin{equation}
\begin{split}
    \frac{\partial^k \sigma (x)}{\partial x^k}=
    \left\{
    \begin{array}{lc}
    1, & \text{if }k=1 \\
    0. & \text{else} \qquad 
    \end{array}
    \right.
\end{split}
\end{equation}

\section{Convergence of the Taylor series}
\subsection{A simple analyze of the convergence}
When the activation function is infinitely differentiable, we can calculate all the derivatives and thus the neural network is equivalent to its Taylor series. 

For unmixed partial derivatives, from Eqns.~(\ref{vMv app})(\ref{beta app}), the high-order derivatives are related to ${\left({\mathbf{W}^{(m+1)}}^T\right)}^{\circ k}$. For mixed partial derivatives, from Eq.~(\ref{total derivatives}), $\frac{\gamma^k \mathbf{y}}{\gamma \mathbf{x}^k}$ is related to $\mathbf{W}_k$ and combination of derivatives of activation function. According to Eq.~(\ref{Wk}), $\mathbf{W}_k$ includes $\mathbf{W}_{i_1,j_1}^{(1)}\mathbf{W}_{i_2,j_2}^{(1)}\ldots \mathbf{W}_{i_k,j_k}^{(1)}$, which is the continuous multiplication of $k$ weights in $\mathbf{W}^{(1)}$. 

\begin{equation}
\begin{split}
    |\frac{\partial^k\B y}{\partial\B x^k}|\propto |\mathbf{W}_{i,j}^{(m)}|^k.
\end{split}
\end{equation}

When the parameters in $\mathbf W^{(m)}$ is concentrated near 0, higher-order derivatives are more likely to approach 0. When the parameters is located far from 0, higher-order derivatives may become increasingly larger due to continuous addition and multiplication, and thus the Taylor series diverge, i.e., we cannot obtain the Taylor approximate solution. The same conclusion also applies to Eq.~(\ref{CT matrix}). 

\begin{equation}
\begin{split}
    \lim_{\substack{\mathbf{W}_{i,j}^{(m)} \to 0 \\ k \to \infty}} |\frac{\partial^k\B y}{\partial\B x^k}| = 0, 
    \lim_{\substack{\mathbf{W}_{i,j}^{(m)} >1 \\ k \to \infty}} |\frac{\partial^k\B y}{\partial\B x^k}| = +\infty.
\end{split}
\end{equation}

The above analysis tells that the parameter distribution of each layer has a great influence on the convergence of Taylor expansion. The above rules help imposing constraints on the network parameters during the network training, and can also help designing network structures with high-order Taylor approximation.

\section{Time Complexity Analysis of \M}
The core algorithm of deep learning is back-propagation, and most of the deep learning frameworks adapt automatic differentiation module, like Autograd. Here, we analyze and compare the time complexity of Autograd and \M~ for a $p$-D neural network. 

(i) Autograd calculates derivatives based on computational graphs whose length increase exponentially at base 2. There are $p^k$ $k$-order derivatives, and the length of their computational graphs is $2^{k-1}$. 

The time complexity $T(n)=\sum_{k=1}^{n}p^k2^{k-1} \sim \mathcal{O}((2p)^n)$. 

(ii) \M obtains all the derivatives at one time, with the main calculations lie in calculating the transformation matrix $\B M$ and conducting back propagation. 

$\B M$ is a lower triangular matrix and the block matrices in $k$-th row need $k$ operations, so the complexity of calculating $\B M$ is $T(n)=\sum_{k=1}^{n}k^2=\f{n(n+1)(2n+1)}{6}\sim\mathcal{O}(n^3)$. For linear layers, $\B M$ turns into a diagonal matrix and the complexity reduces to $T(n)=\sum_{k=1}^{n}k=\f{n(n+1)}{2}\sim\mathcal{O}(n^2)$. For mixed partial derivatives, $\B M$ is a diagonal matrix and the size of $\B Q_k$ is $p^{k-1}$ times larger than $\B W$, the complexity is about $T(n)=\sum_{k=1}^{n}p^{k-1}=\f{1-p^n}{1-p}\sim\mathcal{O}(p^n)$. Therefore, the complexity of \M~ $\mathcal{O}(n^2)<T(n)<max(\mathcal{O}(n^3),\mathcal{O}(p^n))$.

\section{Experiments details}

\subsection{1D function} 

The 1D 4th-order Harmonic oscillator system is defined as
\begin{equation}  
\begin{split}
\left\{  
    \begin{array}{ll}
    u_{tttt}+2u_{tt}+u=0, & t\in [0,2\pi], \\
    u(0)=0, u_t(0)=1, u_{tt}(0)=0.
    \end{array}  
\right.  
\end{split}
\end{equation} 
The initial conditions indicate that the initial position of the harmonic oscillator is the balance point, the initial speed is 1, and the initial acceleration is 0. 

We set $\lambda=5$ and $\mu=1$ for the loss function. The model has 5 layer, with 64 unit in hidden layer and Sine activation function. We trained this model for 1000 epochs, with a Adamax optimizer and a learning rate of 1e-3, and the batch size is 1024.

The output of the network on input $\pi$ is 0.0059, and the first 10 order derivatives are -0.9943, 0.0097, 0.9986, -0.0235, -1.0089, 0.0680, 1.0894, -0.2868, 1.9500, 1.5491, from which we can guess that the true solution is $u=sin(t)$. The 10-order Taylor polynomial is 
\begin{equation}
\begin{split}
    f(t)&=0.0059-\f{0.9943}{1!}\Delta_{\pi}+\f{0.0097}{2!}\Delta_{\pi}^2+\f{0.9986}{3!}\Delta_{\pi}^3-\f{0.0235}{4!}\Delta_{\pi}^4-\f{1.0089}{5!}\Delta_{\pi}^5\\
    &+\f{0.0680}{6!}\Delta_{\pi}^6+\f{1.0894}{7!}\Delta_{\pi}^7-\f{0.2868}{8!}\Delta_{\pi}^8-\f{1.9500}{9!}\Delta_{\pi}^9+\f{1.5491}{10!}\Delta_{\pi}^{10},
\end{split}
\end{equation}
where $\Delta_{\pi}=t-\pi$.

\subsection{2D function} 

We solve a fourth-order and an eighth-order 2D partial PDE using \M.
The first one is a 2D 4th-order Biharmonic equation defined over $(x_1,x_2)\in [0,\pi]\times [0,\pi]$, and its PDE condition is

\begin{equation}  
\left\{  
    \begin{array}{ll}
    \nabla^4u=4sin(x_1+x_2), & (x_1,x_2)\in [0,\pi]\times [0,\pi] \\
    u(0,x_2)=sin(x_2), & x_2 \in [0,\pi] \\
    u(pi,x_2)=-sin(x_2), & x_2 \in [0,\pi] \\
    u(x_1,0)=sin(x_1), & x_1 \in [0,\pi] \\
    u(x_1,pi)=-sin(x_1), & x_1 \in [0,\pi] \\
    u_{x_1x_1} = u_{x_2x_2} = -sin(x_1 + x_2), & (x_1,x_2)\in [0,\pi]\times [0,\pi] \\
    \end{array}  
\right.  
\end{equation} 
where $\nabla^4$ is the fourth power of the del operator and the square of the Laplacian operator $\nabla ^{2}$ (or $\Delta$). 

We set $\lambda=1$ and $\mu=1$ for the loss function. The model has 5 layer, with 64 unit in hidden layer and Sine activation function. We trained this model for 1000 epochs, with a Adamax optimizer and a learning rate of 1e-3, and the batch size is 1024.

The second one is a 2D 8th-order Helmholtz equation defined over $(x_1,x_2)\in [0,1]\times [0,1]$, and its PDE condition is

\begin{equation}  
\left\{  
    \begin{array}{ll}
    \Delta^4u+u=17e^{-x_1-x_2}, & (x_1,x_2)\in [0,1]\times [0,1] \\
    u(0,x_2)=e^{-x_2}, & x_2 \in [0,1] \\
    u(1,x_2)=e^{-x_2-1}, & x_2 \in [0,1] \\
    u(x_1,0)=e^{-x_1}, & x_1 \in [0,1] \\
    u(x_1,1)=e^{-x_1-1}, & x_1 \in [0,1] \\
    \end{array}  
\right.  
\end{equation} 

We set $\lambda=1$ and $\mu=1$ for the loss function. The model has 5 layer, with 64 unit in hidden layer and Sine activation function. We trained this model for 1000 epochs, with a Adamax optimizer and a learning rate of 5e-3, and the batch size is 1024.

\subsection{3D function} 

Here we use \M~ to solve the 4th-order PDE of a heat equation defined over $(t,x_1,x_2)\in [0,4]\times [0,1] \times [0,1]$.

\begin{equation}
\begin{split}
\left\{
    \begin{array}{ll}
        u_t-\nabla^4u=\pi^2 sin(\pi x_1)sin(\pi x_2)(cos(\pi t)-4\pi^2 sin(\pi t)), & (t,x_1,x_2) \in [0,4]\times [0,1]\times [0,1] \\
        u(0,x_1,x_2)=0, & (x_1,x_2) \in [0,1]\times [0,1] \\
        u(t,0,x_2)=u(t,1,x_2)=0, & (t,x_2) \in [0,1]\times [0,1] \\
        u(t,x_1,0)=u(t,x_1,1)=0. & (t,x_1) \in [0,1]\times [0,1] 
    \end{array}
\right.
\end{split}
\end{equation}

We set $\lambda=1$ and $\mu=1$ for the loss function. The model has 7 layer, with 64 unit in hidden layer and Sine activation function. We trained this model for 1000 epochs, with a Adamax optimizer and a learning rate of 1e-3, and the batch size is 256. In this experiment, the parameters of network are limited from -0.9 to 0.9 during training.

\end{document}